%% file: 00-main-arxiv-v2.tex
\documentclass[sigplan,screen,nonacm]{acmart}
\usepackage[labelfont=bf,textfont=normalfont]{caption}
\usepackage[labelfont=bf,textfont=normalfont]{subcaption}
\setcitestyle{round}
\usepackage{hyperref}

\usepackage{draftwatermark}
\SetWatermarkText{\textit{preprint in submission}}
\SetWatermarkScale{.35}
\SetWatermarkColor{red!20!blue!10!white}


\begin{abstract}
\input{sections/00-abstract-v3}
\end{abstract}

\usepackage{xspace}
\usepackage{mfirstuc}

\newcommand\hmm[1]{\ifnum\ifhmode\spacefactor\else2000\fi>1005 \uppercase{#1}\else#1\fi}
\newcommand{\hippocorpus}{\textsc{Hippocorpus}\xspace}
\newcommand{\bagChainName}{sequentiality\xspace}
\newcommand{\BagChainName}{Sequentiality\xspace}

\newcommand{\timeSinceEvent}{\textsc{timeSinceEvent}\xspace}
\newcommand{\freqOfRecall}{\textsc{freqOfRecall}\xspace}

\newcommand{\dropcap}[1]{#1}

\newcommand{\newText}[1]{#1}
\newcommand{\oldText}[1]{}

\begin{document}

\title{Imagined versus Remembered Stories: Quantifying Differences in Narrative Flow}

\author{Maarten Sap}
\authornote{Equal contribution}
\email{msap@cs.washington.edu}
\affiliation{%
  \institution{University of Washington\\Allen Institute for AI}
  \city{Seattle}\state{WA}\country{USA}
}

\author{Anna Jafarpour}
\authornotemark[1]
\email{annaja@uw.edu}
\affiliation{%
  \institution{University of Washington}
  \city{Seattle}\state{WA}\country{USA}
}

\author{Yejin Choi}
\email{yejin@cs.washington.edu}
\affiliation{%
  \institution{University of Washington\\Allen Institute for AI}
  \city{Seattle}\state{WA}\country{USA}
}

\author{Noah A. Smith}
\email{nasmith@cs.washington.edu}
\affiliation{%
  \institution{University of Washington\\Allen Institute for AI}
  \city{Seattle}\state{WA}\country{USA}
}

\author{James W. Pennebaker}
\email{pennebaker@utexas.edu}
\affiliation{%
  \institution{University of Texas at Austin}
  \city{Austin}\state{TX}\country{USA}
}

\author{Eric Horvitz}
\email{horvitz@microsoft.com}
\affiliation{%
  \institution{Microsoft}
  \city{Redmond}\state{WA}\country{USA}
}

\maketitle

\section*{Introduction}
\input{sections/01-00-intro}

\input{figures/two-graphical-models}

\section*{\BagChainName with Large-scale Language Models}
\input{sections/02-01-computational-methods}

\input{figures/hc_figure}
\section*{Results}
\input{sections/03-00-findings}

\input{figures/hc_event_link_figure}


\section*{Discussion}
\input{sections/05-discussion}


\section*{Materials and Methods}
\input{sections/06-methods}
\section*{Acknowledgements}
Microsoft provided an internship for M.A.S, support for crowdworkers, and computing resources for running the GTP-3 neural language model on Hippocorpus. A.J. was supported by NIH Brain Initiative grant K99MH120048. We thank Paul Koch for his assistance with GPT-3 compute.
We also thank Zhilin Wang and members of the Buffalo Lab at the University of Washington for valuable discussions.

\bibliographystyle{ACM-Reference-Format}
\bibliography{00-references,00-ACL-paper-refs}

\clearpage
\appendix 
\onecolumn

\input{sections/10-appendix.tex}
\end{document}

%% file: sections/00-abstract-v3.tex
Lifelong experiences and learned knowledge lead to shared expectations about how common situations tend to unfold. 
Such knowledge of narrative event flow enables people to weave together a story.
However, comparable computational tools to evaluate the flow of events in narratives are limited. 
We quantify the differences between autobiographical and imagined stories by introducing \textit{\bagChainName}, a measure of narrative flow of events, drawing probabilistic inferences from a cutting-edge large language model (GPT-3).
\BagChainName captures the flow of a narrative by comparing the probability of a sentence with and without its preceding story context.
We applied our measure to study thousands of diary-like stories, collected from crowdworkers about either a recent remembered experience or an imagined story on the same topic.
The results show that imagined stories have higher \bagChainName than autobiographical stories and that the \bagChainName of autobiographical stories increases when the memories are retold several months later. 
In pursuit of deeper understandings of how \bagChainName measures the flow of narratives, we explore proportions of major and minor events in story sentences, as annotated by crowdworkers. 
We find that lower \bagChainName is associated with higher proportions of major events.
The methods and results highlight opportunities to use cutting-edge computational analyses, such as \bagChainName, on 
large corpora of matched imagined and autobiographical stories to investigate the influences of memory and reasoning on language generation processes.

%% file: sections/01-00-intro.tex

\dropcap{W}hen we tell a story, we weave together sets of events to form a coherent narrative \citep{bartlett1932remembering, Kurby2008-km, Black1981-kb}.
The narrative flow of those events is influenced by our recollection of experiences from episodic memory \cite{Conway1996-nx,Conway2003Neurophysiological,tulving1972episodic} as well as common knowledge about prototypical sequences of events, 
referred to as {\em schema}  \cite{bartlett1932remembering,schank1977scripts,kintsch1988role,Graesser1981-uj,bower1979scripts,hyman1998errors}.
For example, telling an \textit{imagined} story about a friend's wedding relies on common knowledge about the schema of how a wedding in their culture unfolds.
In contrast, a recalled story drawn from memories about a friend's wedding involves an \textit{autobiographical} recollection of episodic details about experienced events in addition to the knowledge of wedding schema \cite[][]{gilboa2018autobiographical}.
Furthermore, in autobiographical stories, the extent to which schema and episodic details are used in storytelling changes with time passing, as
memories of 
experience become consolidated and schematized into more abstract, semantic, and ``gist-like'' versions \cite{Van_Kesteren2012-el,smorti2016narrating,clewett2019transcending}.

\newText{
A key element of narrative storytelling is referencing occurrences of \textit{salient} events \cite{Sims2019Literary}, which often deviate from prototypical schema \cite{zwaan1998situation}. 
Such salient events can range from major (e.g., big plot twists) to minor (e.g., subtle details) \cite{jafarpour2019medial}, and from surprising to expected.
Small-scale human studies have demonstrated that salient events often mark surprising or expected shifts in a story \cite[e.g., of character focus, location, or circumstances;][]{zacks2007event}, that they stand out as particularly memorable to readers \cite{reichardt2020novelty,franklin2020structured}, and they can influence the emotional impact of a narrative \cite{nabi2015role}. 
However, how salient events contribute to the narrative flow of imagined or autobiographical stories is not well understood.
}

\newText{
We introduce a novel computational measure, \textit{\bagChainName}, to uncover how autobiographical and imagined stories differ with respect to narrative flow and occurrences of salient events.
\BagChainName leverages probabilities of words and sentences in stories to determine the difference in the likelihood of story sentences conditioned on a story's topic versus conditioned on the story topic and the context given by all of the preceding sentences (Fig.~\ref{fig:graph-model}).
In this work, we draw probabilities from a cutting-edge and large-scale language model \citep[GPT-3, with 175B parameters;][]{brown2020language}, substantially scaling up our previous investigations \cite{sap-horvitz-etal-2020-recall-vs-imagined} that employed a much smaller language model \citep[GPT-1, with 115M parameters;][]{Radford2018gpt}.
By using large-scale language models, \bagChainName presents a novel characterization of narrative flow in stories, contrasting with previous measures which either focused on detecting event words from sentences \cite{li2013story,Sims2019Literary} or tracking attributes over time in stories \cite[e.g., sentiment, emotion, categories of words, or sentence embeddings;][]{reagan2016emotional,Boyd2020-jq,Toubia2021-fm}.
}


\newText{
We studied \bagChainName and salient events in a set of 7,000 diary-like short stories about memorable life experiences, to analyze differences in narrative flow of imagined or autobiographical stories.
Collected through crowdsourcing and made available in the \hippocorpus data set \cite{sap-horvitz-etal-2020-recall-vs-imagined}, these stories were either written about an autobiographical personal experience, \textit{recalled} shortly after it happened and \textit{retold} several months later, or about an \textit{imagined} experience on the same topic.
We extended a subset of 240 \hippocorpus stories to additionally include sentence-level human annotation of event saliency.
We applied \bagChainName to these stories to analyze narrative flow difference in autobiographical and imagined stories, and to compare the \bagChainName of sentences with various levels of event saliency.
We also analyzed coarser-grained metrics such as counting concrete and non-hypothesized event terms (i.e., \textit{the proportion of realis event}) and counting words in the LIWC \cite{tausczik2010psychological} and concreteness \cite{brysbaert2014concreteness} lexicons to further examine the differences in stories and event types.

}

We hypothesized that autobiographical and imagined stories would differ in \bagChainName and event distributions; specifically, that imagined stories would have higher \bagChainName since they are more likely to follow commonly expected schema \cite{Conway2003Neurophysiological,Kourken_Michaelian2018-lg}. 
On the other hand, we hypothesized that autobiographical stories would have lower \bagChainName but higher number of annotated salient events, based on the intuition that those stories likely contain more specific details drawn directly from episodic memory \cite{Greenberg1996-yt,Conway2003Neurophysiological} and that memorable details of a specific experience are more likely to diverge from the expected flow of the narrative \cite{reichardt2020novelty}.
We also expected to find an increase in \bagChainName for stories that are retold after a period of time versus freshly recalled memories, due to the consolidation and narrativization of memories over time \cite{smorti2016narrating}.

%% file: figures/two-graphical-models.tex
\begin{figure}[t]
    \centering
    \includegraphics[width=1\columnwidth]{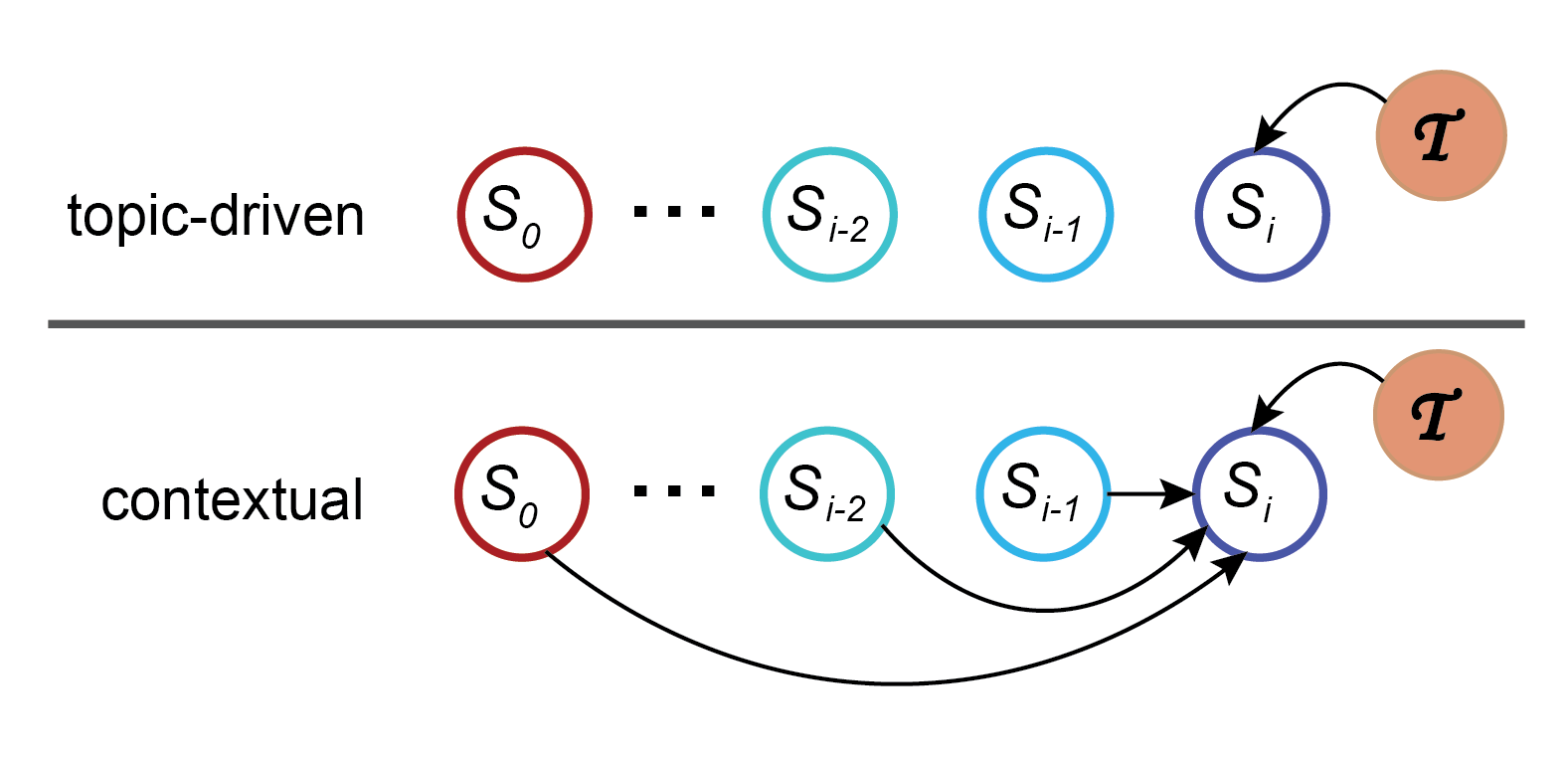}
    \caption{\textbf{Graphical models depicting the two components of \bagChainName.} \BagChainName reflects the probabilistic relationship among consecutive sentences ($s_0, s_1, ..., s_i$) in a story about a topic $\mathcal{T}$, and 
    is computed as the difference between the log-likelihood of a sentence conditioned only on the story topic (i.e., \textit{topic-driven} model; top row) and the log-likelihood of that sentence conditioned on both the story topic and all preceding sentences (i.e., \textit{contextual} model; bottom row). The log-likelihood of a sentence given a topic or topic and prior sentences is provided by the GPT-3 neural language model.
    }
    \label{fig:graph-model}
\end{figure}

%% file: sections/02-01-computational-methods.tex

\newText{{\BagChainName} provides a measure of narrative flow based on probabilities of story sentences given by large language models (LLMs).  
We apply the measure to identify differences in the sequencing of ideas in recalled versus imagined stories. 
One might expect that imagined stories composed in real-time would tend to be described by a contextual model where a next sentence depends greatly on the prior sentences, with a sequencing influenced by commonly understood schemas \cite{Conway2003Neurophysiological,Kourken_Michaelian2018-lg}. 
In contrast, generating an autobiographical story may rely less on such an incremental sequencing and prototypical schemas \cite{reichardt2020novelty} and be better explained by a process of organizing and building a narrative from a set of events encoded in episodic memory \cite{Greenberg1996-yt,Conway2003Neurophysiological}. 

The \bagChainName metric compares, for all sentences of a story, the differences in likelihood for each sentence as predicted by a \textit{contextual} sequencing model versus as predicted by a \textit{topic-driven} model where each sentence is conditioned only on the topic. 
That is, given sentences from a story written about a topic $\mathcal{T}$, \bagChainName compares the likelihood of each story sentence under two generative models, illustrated in Fig. \ref{fig:graph-model}.
The \textit{contextual} generative model assumes that each sentence is generated conditioned on the story topic as well as all of its preceding sentences. 
The \textit{topic-driven} generative model assumes that every generated sentence is  conditioned only on the story topic.
As such, higher values of \bagChainName for sentences suggests that the sentences follow the common expectations given the context of the evolving story and topic, whereas lower values suggest that sentences deviate more from expectation, given the preceding sequence of sentences.
Here, we first briefly introduce LLMs, then formally define \bagChainName, and finally discuss word-based narrative measures that we also use in our experiments.  
}

\subsection*{Large-scale language models}
\newText{
LLMs are a new family of language models (LMs) represented as large-scale neural networks, which have rapidly come to serve as the foundation of most current NLP systems \cite{bommasani2021opportunities}.
Formally, a language model is a statistical model that estimates the \textit{likelihood} or probability of sequences of words, i.e., one or more sentences.
We denote this likelihood as $p_\text{LM}(s_{0:i})$ where $s_{0:i} = \{s_0,..., s_i\}$ are consecutive sentences.
LLMs are trained to estimate the likelihoods of sentences using massive amounts of text.
For example, the model we use in our experiments \cite[GPT-3;][]{brown2020language} is a 175B parameter neural LM trained on over 45TB of text data (e.g., books, news articles, Wikipedia pages).
Through training on such large amounts of text, LLMs also learn an estimate of the general ordering or expected narrative flow of events and sentences in stories \cite{laban-etal-2021-transformer,clark-etal-2021-thats}.
}




\subsection*{Formalization}
\newText{\BagChainName $c(s_i,h)$ is measured for each sentence $s_i$ of a story about topic $\mathcal{T}$ for a number $h$ of preceding sentences (the history under consideration, $s_{i-h:i-1}$). 
$c(s_i,h)$ is computed for each sentence $s_i$, as the difference in the negative log-likelihood (NLL) of the sentence, as computed by the contextual and topic-driven models. 
This requires computing the likelihood of each sentence, conditioned on $h$ prior sentences, per the history $s_{i-h:i-1}$ under consideration, in addition to the words in the story topic $\mathcal{T}$ and, separately, computing the likelihood of the same sentences when each is conditioned only on the story topic $\mathcal{T}$:

$$
    c(s_i,h) = -\frac{1}{|s_i|}[ \log \underbrace{p_\text{LM}(s_i \mid \mathcal{T})}_{\text{topic-driven}} - \log \underbrace{
    p_\text{LM}(s_i\mid \mathcal{T}, s_{i-h:i-1})}_{\text{contextual}} ]
$$
where we obtain the likelihood of sentences $p_\text{LM}$ from LLMs, and normalize the likelihoods by sentence length $|s_i|$ to account for sentence-length variation.
We then define the overall \bagChainName of the entire story as the average \bagChainName of its sentences.

In our analyses, we examined the average \bagChainName per story for history sizes ranging from one to nine preceding sentences ($h=1,...,9$) to the full preceding history ($h=\text{full}$).
We use the story summaries written by the storytellers as the topic $\mathcal{T}$.
We compare \bagChainName to the topic-driven likelihood of sentences, computed by conditioning the sentences of stories only on the topic; we report the negative log-likelihood of sentences, NLL$_\mathcal{T}=-\frac{1}{|s_i|}\log p_\text{LM}(s_i \mid \mathcal{T})$. 
We also compare to the fully contextual negative log-likelihood: NLL$_\mathcal{C}=-\frac{1}{|s_i|}\log p_\text{LM}(s_i \mid \mathcal{T}, s_{0:i-1})$.
}

\subsection*{Lexicon-centric measures}
\newText{
In addition to \bagChainName, we examined the events in narratives and employed several word-based metrics to analyze narratives. The latter lexicon-centric measures,
include counts of the prevalence of \textit{realis} event words, i.e., non-hypothetical references to concrete events that took place (e.g., ``she \textit{tripped}'') in contrast to hypothesized events (e.g., ``she feared \textit{tripping}'', but she did not trip.).
To find those words, we used an automated tagger trained on an annotated corpus of realis terms \cite{Sims2019Literary}.
We also noted average numbers of words in stories falling in psychologically related categories using the Linguistic Inquiry Word Count \cite[LIWC;][]{tausczik2010psychological} lexicon, and measured the average concreteness level of words using a concreteness lexicon \cite{brysbaert2014concreteness}.
To ensure the validity of the concepts measured by these lexicon-based measures \cite{Jaidka2020-gz}, we show the most frequent words in each lexicon category along with our results.
}

%% file: figures/hc_figure.tex
\begin{figure*}[t]
    \centering
    \includegraphics[width=1.5\columnwidth]{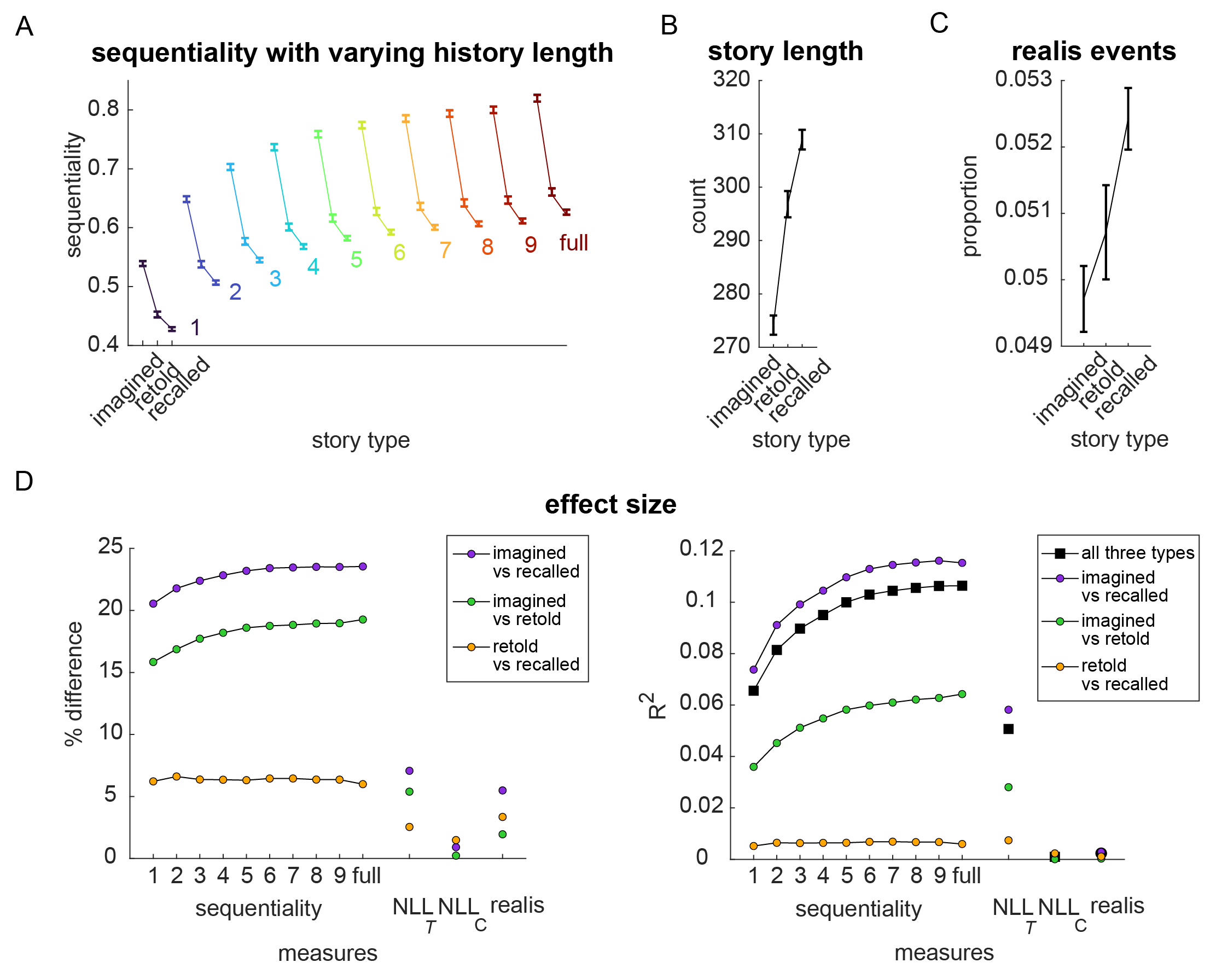} 
    \caption{\textbf{Differences in \bagChainName in recalled, retold, and imagined stories.} 
    (A) Mean \bagChainName of stories with varying history lengths ($h=1$ to h = full story length) are different across the story types. Imagined stories have higher average \bagChainName than autobiographical stories, and retold stories more \bagChainName than recalled stories.
    (B) Stories about imagined events are shorter than autobiographical stories.
    (C) Proportion of realis events is higher in autobiographical stories than in imagined stories. 
    (D) Effect sizes: Percentage difference in parameter estimates (left) and $R^2$ (right), reflecting the magnitude of difference in \bagChainName, the total number of words (story length), the topic-driven and contextual likelihoods of sentences (NLL$_\mathcal{T}$ and NLL$_\mathcal{C}$) and the proportion of realis across story types.}
    \label{fig:hc-figure}
\end{figure*}

%% file: sections/03-00-findings.tex
\subsection*{Analysis of Hippocorpus stories}
We determined the difference in \bagChainName across the three story types (recalled, retold, and imagined stories) using a factorial linear regression with the story type as the grouping factor and the story length.
We included the story length because recalled stories are longer than retold stories ($p=0.001$), and retold stories are longer than imagined stories ($p<0.001$; Fig.~\ref{fig:hc-figure}C).
We report the $R^2$, which quantifies the proportion of variance in the data that is explained by the group difference, the effect size, and the $p$-values after correction for multiple comparisons using the Bonferroni method.  

\subsubsection*{Imagined stories flow in a more expected manner than autobiographical stories}
The comparisons between the \bagChainName across story types ($N$ = 6854 stories on $N$ = 2788 unique topics) show that imagined stories have higher \bagChainName than autobiographical memories ($p<0.001$ for the main effect of the story type on all \bagChainName history lengths; see Fig.\ref{fig:hc-figure} for the effect sizes).
The pairwise comparisons demonstrate that imagined stories have higher \bagChainName than both retold ($p<0.001$) and recalled ($p<0.001$) stories.
\newText{While there were no differences in contextual likelihood (NLL$_\mathcal{C}$) between story types, 
we observe lower topic-driven likelihood (i.e., higher NLL$_\mathcal{T}$) for sentences of imagined stories versus autobiographical stories.
This suggests that the sentences of imagined stories on average have weaker links to the topic than sentences of autobiographical stories, despite both types of sentences having strong links to the preceding sentences.
}
\newText{However, in general, \bagChainName (with increasing history size) has much larger effect sizes and $R^2$ compared to the likelihood or realis metrics (Fig.~\ref{fig:hc-figure}D), which shows that \bagChainName is a better measure for capturing differences in the narratives of imagined and autobiographical stories.}
\oldText{However, the differences across story types measured with \bagChainName (with increasing history size) dominate differences detected with the overall increased topic-centricity of autobiographical stories with the differences in story length and the number of realis events have lower effect sizes and $R^2$ (Fig.~\ref{fig:hc-figure}D).}





\subsubsection*{Retold autobiographical stories have higher \bagChainName than fresh recollections}
In comparison to freshly recalled stories, stories retold after several months have higher \bagChainName ($p<0.001$), are shorter ($p<0.001$), and contain fewer realis events ($p<0.001$; Fig.~\ref{fig:hc-figure}).
This finding demonstrates systematic shifts in the narratives of autobiographical stories with time, posing questions and framing future research on the consolidation of memories in and influences of such processes on recollection. 
We found that participants' assessments of the frequency of recalling or retelling autobiographical stories 
is not associated with \bagChainName but that \bagChainName is negatively correlated with the number of realis events in stories ($r=-0.08$, $p<0.001$). 

\subsubsection*{Autobiographical stories contain more realis events and concrete and time-and-space words than imagined stories}
We found that the proportion of realis events is higher in recalled autobiographical stories than in imagined stories ($p=0.001$; Fig.~\ref{fig:hc-figure}B), but did not differ when comparing recalled and retold ($p>0.1$) or retold and imagined ($p>0.1$) stories.
The proportion of concrete words, measured with LIWC and concreteness lexicons \cite{brysbaert2014concreteness,liwc2015}, is different across story types ($p<0.001$; supplementary Tab.~\ref{tab:hpc_lwic}) with fewer concrete words being used in imagined versus autobiographical stories (recall: $p<0.001$; retold: $p<0.001$). The proportion of concrete words is not different between recalled and retold stories ($p>0.1$).
Additionally, we found that recalled and retold stories contain greater proportions of words related to cognitive processes, time, space, and motion ($p<0.001$; supplementary Table.~\ref{tab:hpc_lwic}).

\subsection*{Event-annotated subset}
Next, we review the differences in the proportion of salient events in a subset of the \hippocorpus that consists of 240 stories on 80 different topics across the three story types. 
Each story sentence was annotated by eight crowdworkers for whether a sentence expressed a major or minor event, and whether the identified event was expected versus surprising.
To control for the variability in schematic knowledge and subjective understanding of what constitutes a major or minor event, the same groups of eight people annotated sentences from the three stories (imagined, recalled, retold) on each topic. 
We summarized the annotations based on majority voting and evaluated the difference in the proportion of major and minor events in the stories across the three story types using ANOVA including consideration of sentence length (sentences with major events are significantly longer than those with no events or with minor events; $p < 0.001$; Fig.~\ref{fig:hc-event-link-figure}B).
Then we studied the relationships among event annotation and \bagChainName, LIWC, and concreteness lexicons at the sentence level.

\input{figures/three_figure}

\subsubsection*{Autobiographical stories contain more salient events than imagined stories}
We observed a main effect for story type on minor events and expected salient events, but not on major events or surprising events (Fig.~\ref{fig:three-figure}). Specifically, higher proportions of sentences in recalled and retold stories were annotated as minor events ($p=0.007$) and expected events ($p=0.025$) as compared to events in imagined stories.
We found no significant difference in the number of minor, major, expected, or surprising events ($p>0.1$) between recalled stories and their retold versions.

\subsubsection*{Sentences with salient events have lower \bagChainName}
We examined the effect of event type (major, minor, or no event) on the \bagChainName of sentences, similar to how we analyzed the effect on story types.
\BagChainName with any history length show a significant main effect of event type ($p<0.001$; Fig.~\ref{fig:hc-event-link-figure}A).
The sentences marked as containing major events have lower \bagChainName than those with no events ($p < 0.001$, all history lengths; no difference with the minor events, $p>0.1$).
Whereas, sentences with minor events have lower \bagChainName than sentences with no events ($p<0.05$) only when the \bagChainName is measured considering the previous sentence (h = 1) but not with longer history ($h>1$, $p>0.1$). 
The results provide evidence that major events have more global influence in a story than minor events.

\subsubsection*{Sentences with salient events have a higher proportions of realis event terms and concrete, present-related, and space-related words}
We found a higher proportion of realis event term in sentences with minor events than in those with a major ($p<0.001$) or no events ($p < 0.001$; Fig.~\ref{fig:hc-event-link-figure}C).
\newText{Using the LIWC and concreteness lexicons, we generally observe more differences between sentences with no event and those with a salient event, compared to between sentences with a minor vs. major event (see supplementary Table \ref{tab:eventSeg_liwc}).
Notably, in addition to lower proportions of concrete and time-and-space words, we see higher proportions of words related to cognitive and affective processes in sentence with no events.
}

\subsubsection*{Sentences with surprising events have lower \bagChainName than those with expected major events}
We found that major events are often annotated as surprising (72\%) rather than expected (28\%), whereas all minor events are annotated as expected.
Sentences annotated as describing major events have a lower \bagChainName when they are noted to be surprising versus  expected ($p<0.001$).
\bagChainName is also lower for expected major events compared to expected minor events ($p<0.001$; the difference increased with increasing history length).
In general, we found that \bagChainName of sentences is not different for surprising and expected sentences ($p>0.05$; the difference decreased with increasing history length; for h = 1, uncorrected $p = 0.014$), suggesting that \bagChainName captures more than the event expectancy.

%% file: figures/three_figure.tex
\begin{figure}[t]
    \centering
    \includegraphics[width=1\columnwidth]{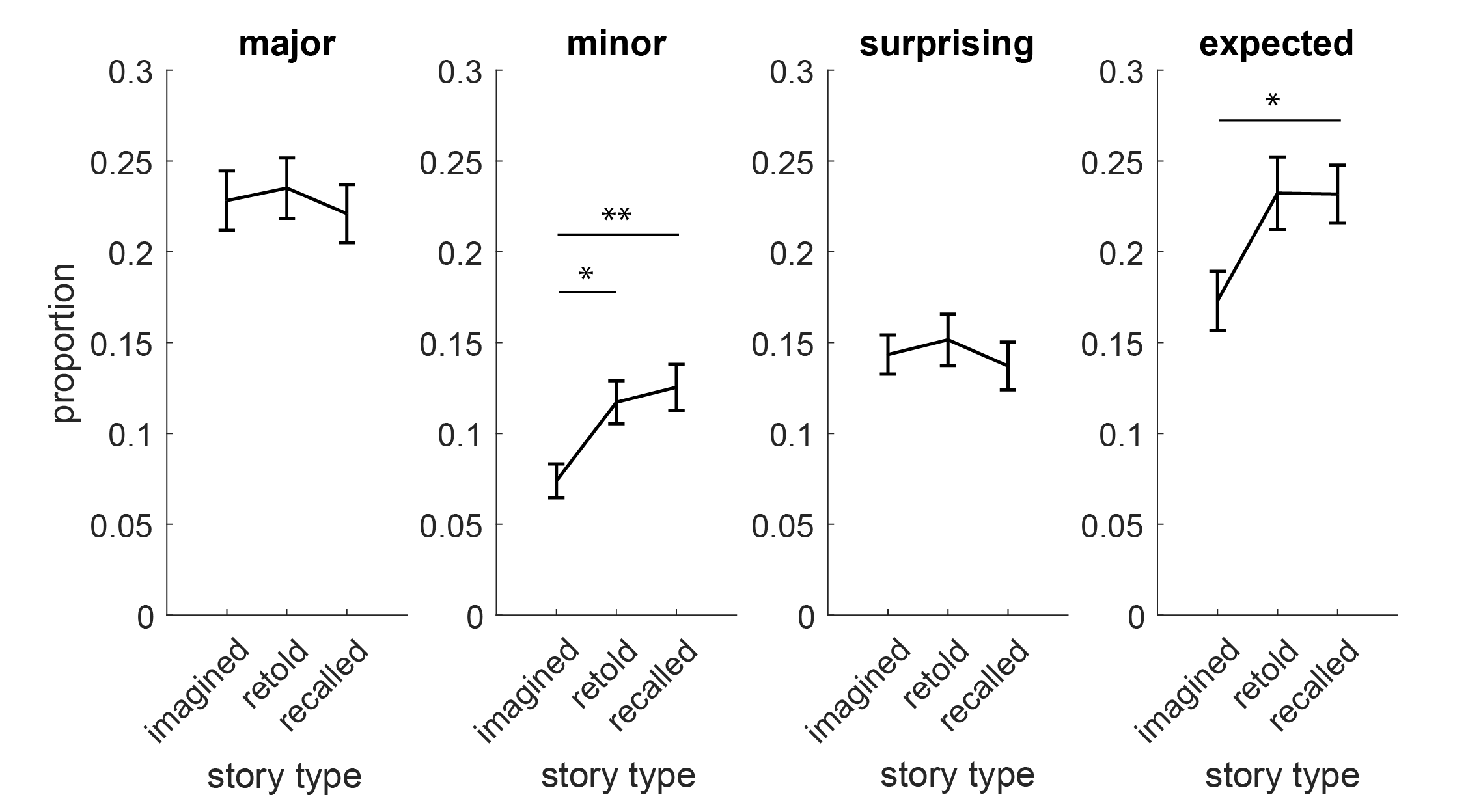}
    \caption{\textbf{Proportions of salient event annotations across the stories}. Graphs of the mean and standard error of the mean of the proportion of events annotated as salient (from left: major, minor, surprising, and expected events) in the imagined, recalled, and retold stories. (* $p<0.05$, ** $p<0.01$)}
    \label{fig:three-figure} 
\end{figure}

%% file: figures/hc_event_link_figure.tex
\begin{figure}
    \centering
    \includegraphics[width=\columnwidth]{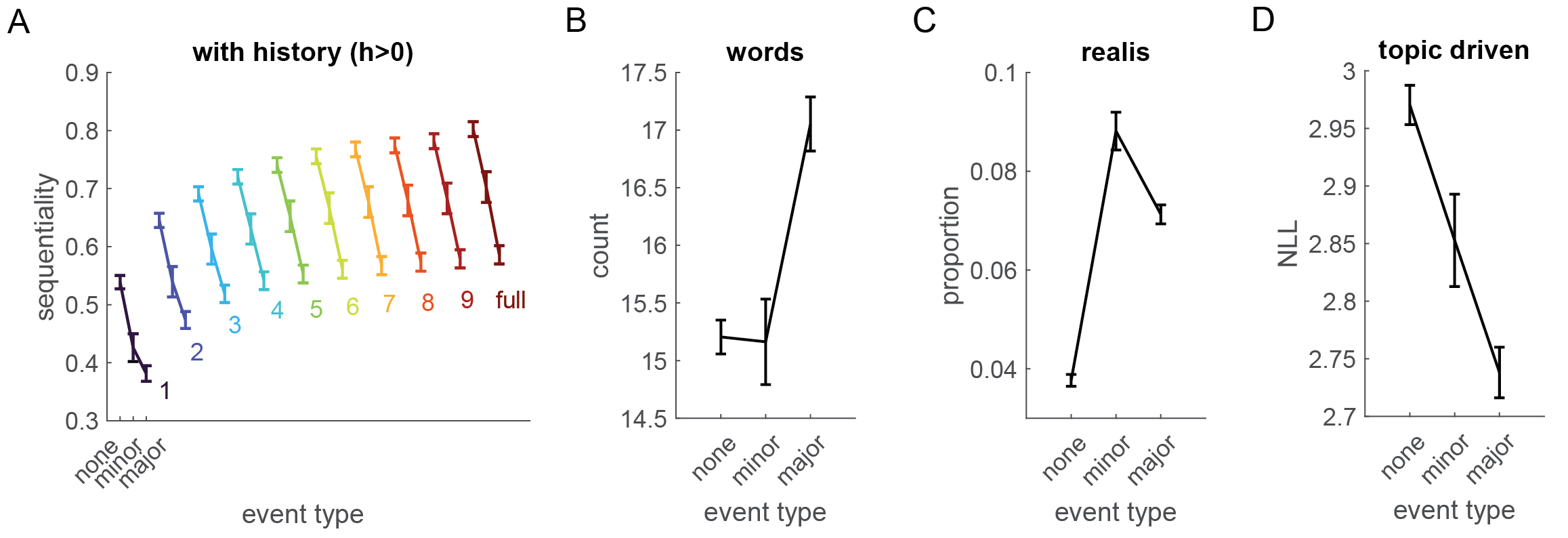}
    \caption{\textbf{\BagChainName of sentences relative to event annotations}. 
    (A) The average \bagChainName, with varying history, is grouped by the event type. The sentences with no event (none) follow the narrative flow of the story topics more than sentences with major (all \bagChainName history length) or minor events do (with \bagChainName history of one sentence). \BagChainName of minor and major events are not different. 
    (B) The sentences with no event are shorter than sentences with major events. 
    (C) The realis in sentences with major or minor events is higher than in sentences with no event. Error bars show standard error of the mean.
    }
    \label{fig:hc-event-link-figure}
\end{figure}

%% file: sections/05-discussion.tex

\newText{We introduced \textit{\bagChainName} as a new computational measure of narrative flow of events instantiated by large-scale neural language models. We used the measure to probe hypotheses about the generative processes of constructing experienced versus imagined stories, paired via a matched topical description starting point.
\BagChainName measures the extent to which story events and sentences flow from their preceding context and overall story topic versus from only the story topic (Fig. \ref{fig:graph-model}), 
using likelihoods given by the GPT-3 large-scale language model \cite{brown2020language}.
As such, \bagChainName can be considered a proxy for quantifying how much a story follows the expected or common narrative flow for a specific story topic (schematic knowledge) versus is grounded in experiential details (episodic memory).

We used \bagChainName to study differences in narrative flow across (1) \textit{recalled} stories based on fresh autobiographical experiences, (2) \textit{retold} stories about those same autobiographical experiences after three to six months, and (3) \textit{imagined} stories matched to the topics of the autobiographical stories.
With \bagChainName and word-based metrics such as the count of realis event terms (that refer to concrete, non-hypothesized event occurrences) and LIWC and concreteness lexicon scores, we observed differences in episodic details and differing reliance on schematic knowledge for constructing narratives.
Based on \bagChainName differences, imagined stories have greater alignment with expected schematic flow of events than autobiographical stories.
Autobiographical stories contain more minor detailed events than imagined stories (Fig. \ref{fig:three-figure}), and they tend to have higher proportions of concrete words as well as words related to time and space (supplementary Table~\ref{tab:hpc_lwic}). 
Below, we discuss implications of our findings 
for analyzing narrative flow of events using large-scale neural LMs, as well as for understanding cognitive processes of storytelling with computational methods.}

\subsection*{Using \bagChainName to quantify narrative flow} 
\newText{\BagChainName is a novel measure to quantify the extent to which the flow of events follows expected schema, using large-scale neural language models.
This is a departure from previous measures of narrative flow, which have predominantly approached the task by examining word usage, such as the rates of emotion-related words over time.
In prior work, researchers have argued that emotional flow plays a role in the persuasiveness of stories \cite{nabi2015role}, an approach which was later operationalized through word-counting of emotion words in books \cite{reagan2016emotional} and consumer reviews \cite{Van_Laer2019-qq}.
In addition to emotion words, recent work computed the progression of the rate of function words and words related to cognitive processes to study narrative progression and their relationship to story quality \cite{Boyd2020-jq}. 
Beyond studying word frequencies, a recent study \cite{Toubia2021-fm} employed high-dimensional word vectors to compute the speed and complexity of stories.
In contrast to previous work which analyzed narratives through surface-level features, 
\bagChainName leverages a story's topic and the language modeling capabilities of large-scale neural LMs to infer the predictability of words. 
\BagChainName does not relying on specific word categories or high-dimensional word vectors.
\BagChainName was initially used to measure the  \textit{linearity} of sentences, in a preliminary investigation \cite{sap2020recollectionImagination} where we used a much smaller neural LM \cite[GPT-1;][]{Radford2018gpt} than the one used in this study.}

\newText{
Conceptually, the \bagChainName generative model provides a new lens on how sentences and events are produced or read, adding to several models of sentence and event processing from cognitive science.
\BagChainName relates to word-level \textit{surprisal theory} \cite{hale2001probabilistic,levy2008expectation}, which posits that humans form expectations of which word should come next in text, before observing it.
Contextual generative models can formalize those expectations (e.g., Fig. \ref{fig:graph-model}; bottom), and neural LMs can approximate these human expectations about words given sufficient context \cite{Goldstein2021thinking}.

However, surprisal theory does not account for the variation in non-contextual likelihood of events depending on the story topic, which may play a role in how humans form expectations.
For example, a story about driving on a highway for 30 miles might have fewer expected events than one about a birthday party, which has opportunities for details on whose birthday it was, where it took place, who attended, how the cake tasted, etc.
We account for this variation by conditioning both the topic-driven and contextual models on the story topic.
Although we find no differences in contextual likelihood (NLL$_\mathcal{C}$) and only small differences in topic-driven likelihoods (NLL$_\mathcal{T}$), the largest difference across story types is measured as the ratio of contextual and topic-driven likelihoods using \bagChainName (Fig.~\ref{fig:hc-figure}D).
Corroborating this need for comparing likelihoods, recent work has shown the usefulness of comparing contextual and non-contextual event likelihoods in a visual event segmentation tasks \cite{franklin2020structured}.
}
\BagChainName is built on the assumption that large-scale neural language models encode knowledge about the commonly expected narrative flow of events.
Previous work suggests that this is a valid assumption, since LLMs can determine the correct ordering of sentence in text \cite{Radford2018gpt,Radford2019gpt2} and can be used to generate expected schemas for events \cite{laban-etal-2021-transformer}.
However, the extent to which LLMs learn the common flow of events is influenced by the knowledge contained in their training data \cite{gordon2013reporting}.
Specifically, the culture and identities of the authors of training data can influence the schema that are deemed likely by the model; a language model trained exclusively on British text only will likely learn British-specific schema (e.g., tea time) that other models might not encode.
However, our findings with \bagChainName remain similar when
using language models trained on other data sets \cite{sap-horvitz-etal-2020-recall-vs-imagined}, such as OpenAI-GPT \citep[trained on 5GBs of English fiction;][]{Radford2018gpt} and GPT-2 \cite[trained on 40GBs of news-like English text;][]{Radford2019gpt2}, suggesting this may not be a substantial issue.

\subsection*{Cognitive processes of recalling versus imagining}

\newText{The results reveal differences in the cognitive processes of how people form narratives grounded in their own experiences versus from their imagination, and the differential role of salient events in both types of storytelling.}  
Although imagination and remembering may engage similar mental processes \cite{Schacter2007-as} and imagination could leverage one's own life experiences \cite{Grysman2011-jj}, we found series of systematic differences between imagined and autobiographical stories. 
In all stories, storytellers appear to combine schematic knowledge with references to major events. 
We found that major events tend to be relayed in surprising sentences that tend to deviate from expectation, per likelihoods provided by the neural language model.
These sentences are associated with the lowest \bagChainName (Fig. \ref{fig:three-figure} and Fig. \ref{fig:hc-event-link-figure}), and they are often about personal concerns and core drives and needs (supp. Table~\ref{tab:eventSeg_liwc}).
For example, in the recalled story on "\textit{A warm summer morning with a hummingbird. How I had a communal moment with nature by misting a hummingbird with a garden hose.}", the major event is that "\textit{At first, I thought he [the hummingbird] was just doing his early morning pollen rituals, but to my surprise he wanted water.}" 
In an imagined story on the same topic, the major event is that "\textit{[animal started to come to the garden.] Mostly squirrels at first and a few deer, and one tiny hummingbird.}"
Similarly in the recalled story, the major event is that "\textit{I saw a hummingbird at the corner of my eye.}"

A significant difference between the autobiographical and imagined stories is in the proportion of minor events, as identified through human annotations (Fig. \ref{fig:three-figure}).
The minor events tend to be non-hypothesized, concrete details of the stories that are noted as expected but typically not part of the general schema of the story topic. The minor events have local saliency and can be identified only with computation of \bagChainName with a one sentence history.
These events often contain words on biological processes and social references.
As an example, a minor event in a recalled story on the same topic as the example above is that "\textit{I was feeling kind of low due to not seeing many of my friends anymore due to everyone being busy with their schedule, and work being a little slow was also on my mind.}" and in an imagined story was that "\textit{For the first few weeks I got nothing and no activity, then about a month ago animals came.}"

Shedding some new light on the nature of salient events, we found that sentences annotated as describing salient events tend to have more concrete words, first-person references, social words, and words related to cognitive processes, biological processes, core drives and needs, and relativity to time, space, and motion.
Only a subset of these observations, including the change in time, character, and space, has been previously reported in studies on detection of salient events to mark an event boundary \cite{kurby2008segmentation}.
We also observed that the length of stories showed small differences among the story types. 
This observation on length is congruent with the understanding that the stories that rely largely on commonly expected schema are generally shorter \cite{thorndyke1977cognitive,kintsch1978role}.

We found that the proportion of salient events (major and minor) are similar in stories about freshly recalled memories and about memories retold after 3--6 months (Fig. \ref{fig:three-figure}). 
The retold stories have higher \bagChainName and are shorter than the initial recall of stories (Fig. \ref{fig:hc-event-link-figure}).
The self-reported frequency of revisiting and retelling autobiographical stories does not appear to influence the \bagChainName of the stories. 
The retold stories that were noted as more frequently revisited memories 
were found to contain fewer realis events, which may reflect processes of abstraction.
The \bagChainName measure provides a means of quantifying the observation that, with passing time and memory consolidation, 
retelling autobiographical memories relies less on recall from episodic memory, 
instead increasingly invokes common semantic knowledge of schema \cite{bartlett1932remembering,bower1979scripts,smorti2016narrating}, especially since certain events may be forgotten \cite{squire1981two}.

\subsection*{Open research directions}
The methods and results presented show promise as tools for exploring  processes of memory, reasoning, and imagination employed to generate narratives. The methods also hold opportunity to help with building deeper understandings of influences of common schema and personal experiences on the stories that people tell.
\newText{From a computational perspective, we see rich opportunities ahead for harnessing large-scale neural models to explore narrative theories, including consideration and comparative study of different generative models \cite{Piper2021-bj}.}
From a cognitive perspective, directions include pursuing answers to standing questions about the contributions of memory and reasoning to the stories that people generate about experienced and imagined events, and how memories---and the autobiographical stories that flow from them---evolve over time since events are experienced. From a cultural perspective, the methods can provide opportunity to study differences across communities and cultures of the nature and influences of common schema and personal experiences on stories. Opportunities for study include seeking insights about the influences and interpretations of world events over time on fiction and non-fiction narrativizations \cite{underwood2013,underwood2020}.
Other research directions include applying the results, methods, and measures in studies of narrativizations with different motivations \cite{Marsh2007-ud} such as recall, storytelling, persuasion, lie-detection, false confessions, recovered memories, and the propagation and effects of misinformation.

%% file: sections/06-methods.tex
\subsection*{Building \hippocorpus}
In our analyses, we make use of our previously collected corpus of autobiographical, imagined, and retold stories \cite[\hippocorpus;][]{sap-horvitz-etal-2020-recall-vs-imagined}.
This corpus contains 6,854 stories collected from crowdworkers in three stages (depicted in Fig.~\ref{fig:crowdsourcing-pipeline}).
In the recalled stage, workers write a short diary-like story and a short summary.
Then, in the imagined stage, workers are given a summary and asked to write a short diary-like story.
Finally, in the retold stage, workers from the first stage are given their original summary and asked to re-tell their story, after 3-6 months have passed.
For both the recalled and retold tasks, we collect from workers the time elapsed since they experienced the event (\timeSinceEvent, in weeks or months), as well as the frequency at which they thought or talked about the event (\freqOfRecall, on a five-point Likert scale of ``never'' to ``constantly'').
For more details, see our preliminary work \cite{sap-horvitz-etal-2020-recall-vs-imagined} and the Supplementary materials \ref{supp:hippocorpus-collection}.

\subsection*{Collecting event annotations}
We additionally collected sentence-level event annotations for a subset of the \hippocorpus stories.
We randomly selected 80 topics and their associated recalled-imagined-retold stories ($N$ = 240 stories).
Since people's individual perceptions of what constitutes an expected, surprising, major, or minor event could differ depending on their experiences, background, or culture, we make sure all stories about the same topic are annotated by the same worker. 
We collected event annotations from 8 crowdworkers per set of three stories.

Participants read each of the sentences in each of the three stories, one sentence at a time, and indicated if the story sentences mark a start of a new event.
Specifically, annotators marked whether a sentence represented a new event that is \textit{minor} or \textit{major} and if the events are \textit{expected} or \textit{unexpected}.
See Supplementary \ref{supp:event-seg} for further details.

\subsection*{Extracting \bagChainName, realis events, and lexicon counts}

To compute the \bagChainName of each story sentence, we first split each story in the \hippocorpus into sentences, using a version of the NLTK sentence tokenizer\footnote{\url{https://www.nltk.org/api/nltk.tokenize.html}} adapted to avoid splitting sentences into one-word sentences.
We then used the OpenAI API to obtain the likelihoods under GPT-3 of each sentence conditioned on the story topic and various history sizes.
Specifically, we compute the log-likelihood of a sentence $p_\text{GPT-3}(s_{i})$ by summing the word-level log-probabilities yielded by the API for the sentence at hand.
We can then compute \bagChainName for each history size.

For computing the proportions of realis event terms, we use a realis term tagger from our preliminary investigations \cite{sap-horvitz-etal-2020-recall-vs-imagined}. 
This tagger is a BERT \cite{Devlin2019bert} model trained on a realis annotated corpus of literary fiction \cite{Sims2019Literary}, which achieves F1 accuracy scores of 83.7\% and
75.8\%, on the validation and test sets, respectively.

We used the LIWC 2015 software\footnote{\url{https://www.liwc.app/}} for counting the proportion of words that belong to specific LIWC categories \cite{liwc2015}.
For the concreteness lexicon \cite{brysbaert2014concreteness}, we averaged the concreteness lexicon of each story by matching words in the story with words in the lexicon.

\subsection*{Data analysis}
For each story, we averaged the \bagChainName of all sentences and had one representative value for each of the \bagChainName with history length of 1 sentence to full story.
We also took the averaged proportion of major or minor events, the averaged proportion of realis events, the total number of words (story length), and the averaged negative log likelihoods (NLL$_\mathcal{C}$, NLL$_\mathcal{T}$) per story.
We applied a factorial linear regression on each of the parameters, to identify the differences between story types. 
We either included three factors for three story types (imagined, retold, recalled) or we included two factors for pairwise comparisons.

We similarly used a factorial linear regression to evaluate the characteristics of sentences with various event type (major, minor, or no events).
A sentence was accepted to be a minor or major event if the majority of the annotators marked the sentence as such. 
We also evaluated the proportion of events that were expected or surprising by the majority of the annotators.
This analysis was done at the sentence level with 9412 major, 6835 minor, and 17477 no event annotation.
We used Bonferroni correction to adjust the significance threshold for multiple comparisons.
All reported p-values are Bonferroni corrected.

%% file: sections/10-appendix.tex


\section{Further Data Collection Details}
\subsection{\hippocorpus}\label{supp:hippocorpus-collection}
\begin{figure}
    \centering
    \includegraphics[width=\textwidth]{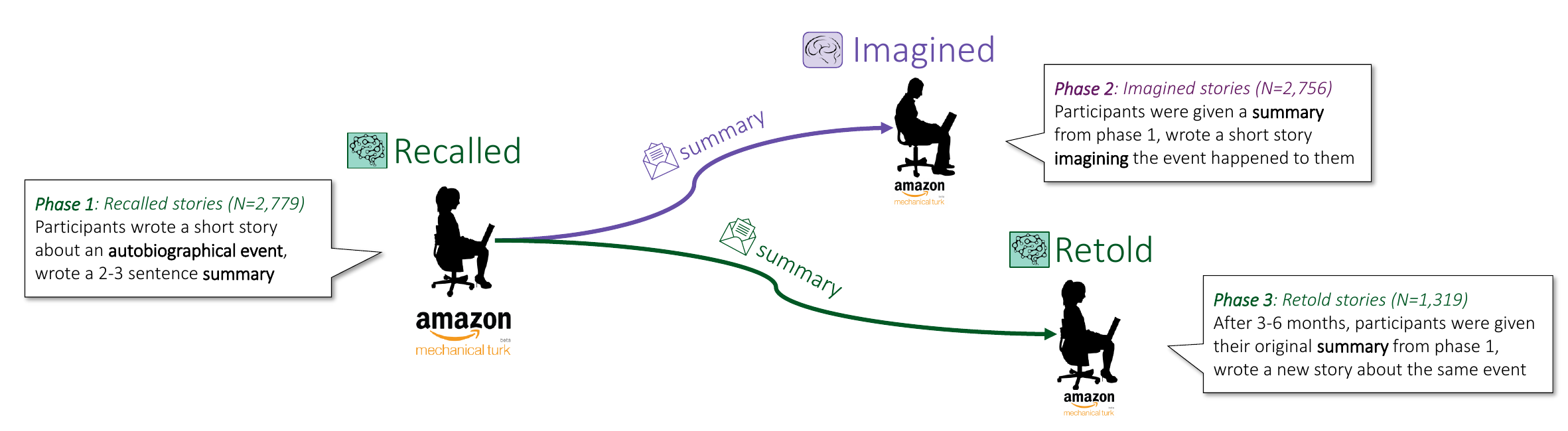}
    \caption{\newText{Data collection pipeline for collecting the \hippocorpus.}}
    \label{fig:crowdsourcing-pipeline}
\end{figure}
The stories in \hippocorpus were collected from crowdworkers in three phases \newText{(see Figure \ref{fig:crowdsourcing-pipeline})}.
In phase one, a set of crowdworkers wrote stories about memorable events they had experienced in the recent past (3-6 months) and summarized their story in one to three sentences ($N$=2,779 \textit{recalled stories}, written by 2662 authors\footnote{2550 of the authors wrote one autobiographical story, 107 wrote two stories, and 5 wrote three stories}).
The summary of the story that was provided by the author was used as the \textit{topic} of the story. 
An example of a topic is: ``\textit{My daughter and her husband announced the were expecting their second child.  While on a camping trip she feared that she might be having a miscarriage only to learn that she was having twins.}''

In the second phase, we used the story topic after 3-6 month of writing the original memory, to ask a subset of the authors to retell the autobiographical story ($N$=1,319 \textit{retold stories}).
In the third phase, we provided the story topics to another set of crowdworkers and asked them to write \textit{imagined stories} as if the event in the summary had happened to them ($N$=2,756 imagined stories, written by $N$=1434 authors\footnote{As participation in this task was not restricted, 1,072 wrote one story, 311 between two and five stories, and 30 between six and ten stories, and 21 workers more than ten stories (following a Zipfian distribution between 11 and 92).}).
During the recalled and retold storywriting tasks, we also asked workers the time elapsed since they experience the event (\timeSinceEvent, in weeks or months), as well as the frequency at which they thought or talked about the event (\freqOfRecall, on a five-point Likert scale of ``never'' to ``constantly'').

\subsection{Event annotations}\label{supp:event-seg}
To test the hypothesis that the \bagChainName of narratives is associated with the number of events contained in stories, we ran an annotation task and analyzed the number of events in a random selection of $N$ = 240 stories of the \hippocorpus.
The subset consisted of a triple of imagination, recalled, and retold stories for each of 80 topics.
In this subset, the autobiographical stories on the same topic are written by the same person, to keep the author's schematic knowledge of the topic constant.
Also, each triple of story types is annotated by the same participants (n = 8).
Keeping the annotators within a topic constant allowed control of the variability in schematic information and the individual difference in event segmentation \citep{jafarpour2019event, jafarpour2019medial}. 

In the annotation task, participants when through three stories, one sentence at a time, and indicated if the story sentences mark a start of a new event.
We specifically asked the annotators to differentiate whether a new event is \textit{minor} or \textit{major} and, for 60 topics, we additionally asked if events are \textit{expected} or \textit{unexpected}. 
Given that the saliency of events can vary \cite{jafarpour2019medial}, participants were instructed to use their interpretation of what constitutes a major or minor event and if the events are expected or surprising. 
The order of the type of story posed for annotation was randomized.

\subsection{Participants} \label{supp:data-demographics}
We recruited a diverse group of story authors. The participants' age ranged between 18 and 55 years old (M = 33.6, SD = 10.5) and were 47\% male, 52\% female, $<$1\% non-binary, and $<$1\% other. 
They were 73.7\% White, 10.1\% Black, 5.2\% Asian, 6.1 \% Hispanic, 0.8\% Native American, 0.7\% Indian, 0.3\% Middle Eastern, 0.2\% Islander, 2.7\% Other, and 0.7 \% unidentified \cite[this data has previously been published in][]{sap-horvitz-etal-2020-recall-vs-imagined}.
189 participants annotated events in a selection of stories (18-55 years old (M=37, SD=10.6); 53\% men, 46\% women, and 1\% unidentified; 75.7\% white, 6.3\% Asian, 5.8\% Black, 4.2\% Hispanic, 0.5\% Indian, 0.5\% Native American, 5.3\% other, 1.6\% unidentified).
All studies were conducted on Amazon Mechanical Turk.
The procedures were approved by Microsoft's ethical review board.
All participants gave written informed consent before participation and were compensated.

\input{tables/LWIC_hippocorpus}

\newcommand{\topic}{\mathcal{T}}





\newText{
\section{Lexicon-based narrative measures: results}
To complement the \bagChainName and realis analyses, we used several lexicon-based measures to investigate the stylistic and content differences in imagined and autobiographical stories, as well as in sentences with minor, major, or no events.
For each story or sentence, LIWC lexicon scores are computed as the proportion of words in that story or sentence that appear in LIWC categories \cite{liwc2015}.
For the weighted concreteness lexicon \cite{brysbaert2014concreteness}, which contains a concreteness rating between 0--1 for 39,000 words, we compute a story's or a sentence's concreteness score as the average rating per word in the story or sentence.

We list the LIWC and concreteness scores, comparing proportions in imagined, recalled, and retold stories (Table \ref{tab:hpc_lwic}) and the proportions in sentences with major, minor, or no events (Table \ref{tab:eventSeg_liwc}).

As a way to verify the validity of these categories in our corpus \cite{Jaidka2020-gz}, we list the top five most frequent words in our two corpora.
Additionally, we also compute the coverage of each category, i.e., the proportion of words in our corpora that belong to each category. 
}

%% file: tables/LWIC_hippocorpus.tex
\begin{table}[t]
    \centering
    \includegraphics[height=.9\textheight]{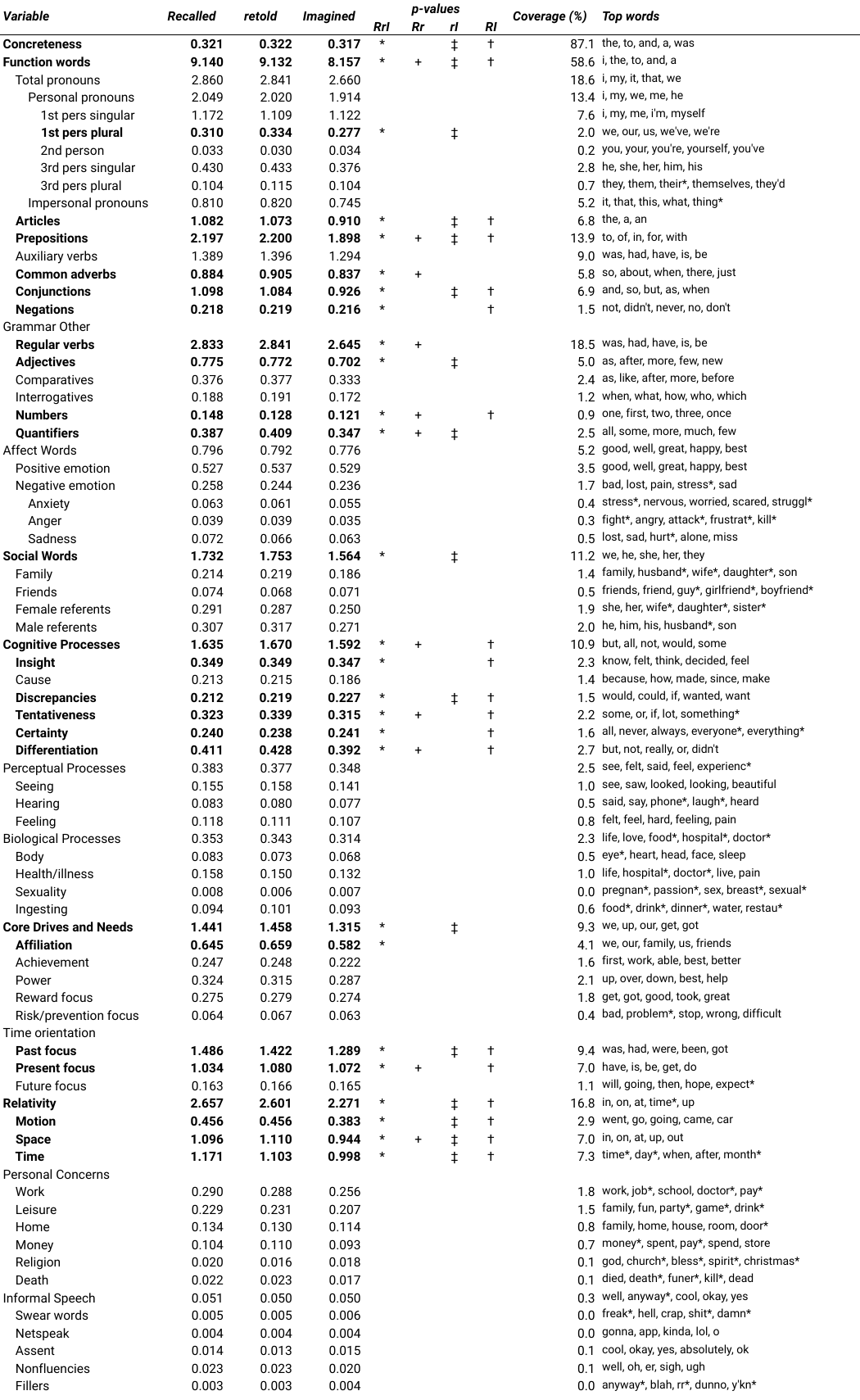}
    \caption{Average lexicon scores for the three story types in \hippocorpus (recalled (R), retold (r), and imagined (I)), along with significance values of the three-way and pairwise differences. The $*$,$+$,$\dagger$,$\ddagger$ symbols denote p-values <0.05 after Bonferroni correction.
    ``Coverage'' indicates the percentage of the total number of words in the \hippocorpus that are in the lexicon category (``Variable''), with the five most common words listed under ``Top words''.
    }
    \label{tab:hpc_lwic}
\end{table}

\begin{table}[t]
    \centering
    \includegraphics[height=.9\textheight]{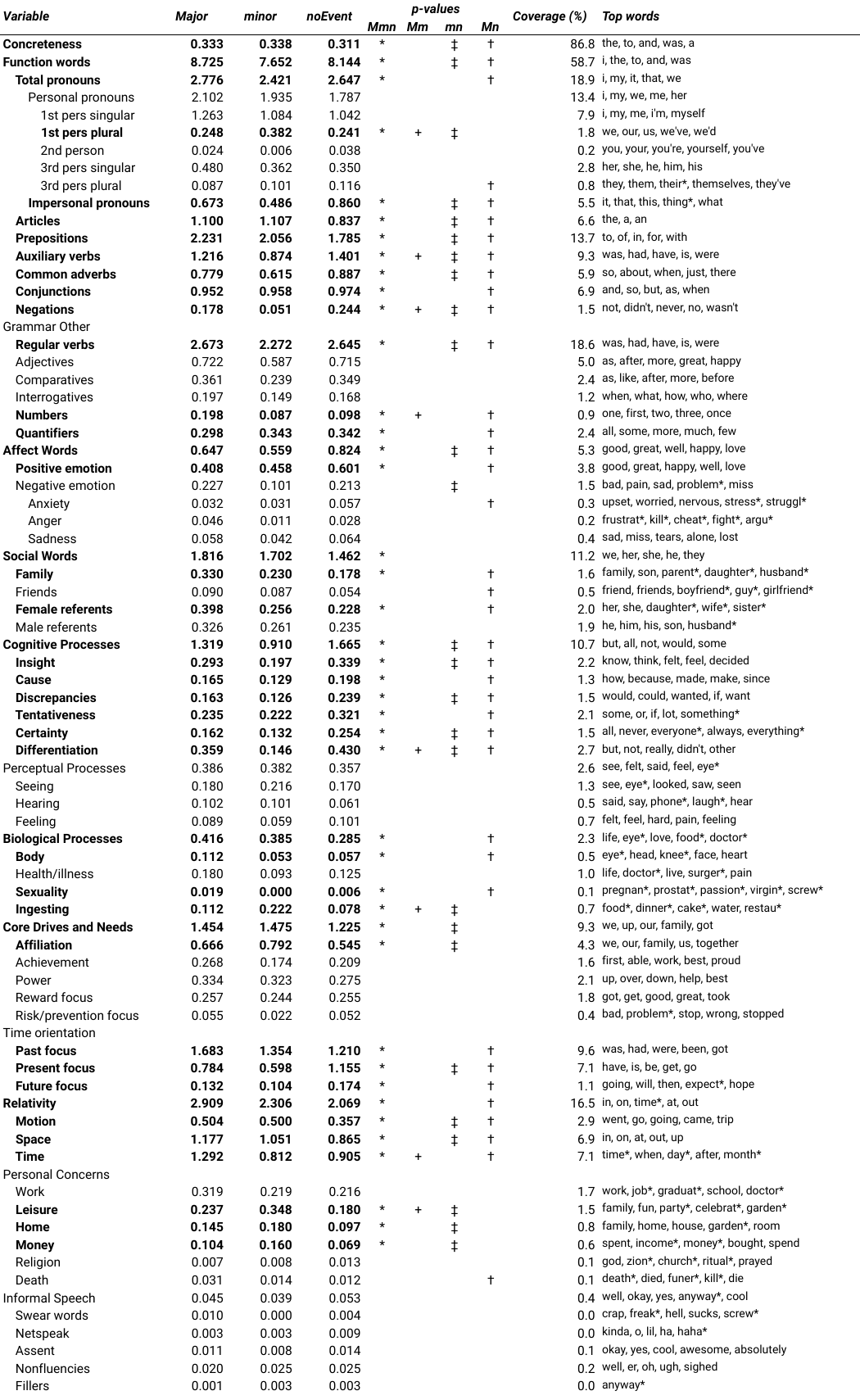}
    \caption{Average lexicon scores for the three event types (major (M), minor (m), and no event (n)) in the annotated subset, along with significance values of the three-way and pairwise differences.  The $*$,$+$,$\dagger$,$\ddagger$ symbols denote p-values <0.05 after Bonferroni correction.
    ``Coverage'' indicates the percentage of the total number of words in the event-annotated sentences that are in the lexicon category (``Variable''), with the five most common words listed under ``Top words''.
    }
    
    \label{tab:eventSeg_liwc}
\end{table}

%% file: 00-main-arxiv-v2.bbl

\begin{thebibliography}{57}


\ifx \showCODEN    \undefined \def \showCODEN     #1{\unskip}     \fi
\ifx \showDOI      \undefined \def \showDOI       #1{#1}\fi
\ifx \showISBNx    \undefined \def \showISBNx     #1{\unskip}     \fi
\ifx \showISBNxiii \undefined \def \showISBNxiii  #1{\unskip}     \fi
\ifx \showISSN     \undefined \def \showISSN      #1{\unskip}     \fi
\ifx \showLCCN     \undefined \def \showLCCN      #1{\unskip}     \fi
\ifx \shownote     \undefined \def \shownote      #1{#1}          \fi
\ifx \showarticletitle \undefined \def \showarticletitle #1{#1}   \fi
\ifx \showURL      \undefined \def \showURL       {\relax}        \fi
\providecommand\bibfield[2]{#2}
\providecommand\bibinfo[2]{#2}
\providecommand\natexlab[1]{#1}
\providecommand\showeprint[2][]{arXiv:#2}

\bibitem[Bartlett(1932)]%
        {bartlett1932remembering}
\bibfield{author}{\bibinfo{person}{Frederic~Charles Bartlett}.}
  \bibinfo{year}{1932}\natexlab{}.
\newblock \bibinfo{booktitle}{\emph{Remembering: A study in experimental and
  social psychology}}.
\newblock \bibinfo{publisher}{Cambridge University Press}.
\newblock


\bibitem[Black and Bern(1981)]%
        {Black1981-kb}
\bibfield{author}{\bibinfo{person}{John~B Black} {and} \bibinfo{person}{Hyman
  Bern}.} \bibinfo{year}{1981}\natexlab{}.
\newblock \showarticletitle{Causal coherence and memory for events in
  narratives}.
\newblock \bibinfo{journal}{\emph{Journal of Verbal Learning and Verbal
  Behavior}} \bibinfo{volume}{20}, \bibinfo{number}{3} (\bibinfo{date}{June}
  \bibinfo{year}{1981}), \bibinfo{pages}{267--275}.
\newblock


\bibitem[Bommasani et~al\mbox{.}(2021)]%
        {bommasani2021opportunities}
\bibfield{author}{\bibinfo{person}{Rishi Bommasani}, \bibinfo{person}{Drew~A
  Hudson}, \bibinfo{person}{Ehsan Adeli}, \bibinfo{person}{Russ Altman},
  \bibinfo{person}{Simran Arora}, \bibinfo{person}{Sydney von Arx},
  \bibinfo{person}{Michael~S Bernstein}, \bibinfo{person}{Jeannette Bohg},
  \bibinfo{person}{Antoine Bosselut}, \bibinfo{person}{Emma Brunskill},
  {et~al\mbox{.}}} \bibinfo{year}{2021}\natexlab{}.
\newblock \bibinfo{title}{On the opportunities and risks of foundation models}.
   (\bibinfo{year}{2021}).
\newblock
\newblock
\shownote{arXiv preprint arXiv:2108.07258}.


\bibitem[Bower et~al\mbox{.}(1979)]%
        {bower1979scripts}
\bibfield{author}{\bibinfo{person}{Gordon~H Bower}, \bibinfo{person}{John~B
  Black}, {and} \bibinfo{person}{Terrence~J Turner}.}
  \bibinfo{year}{1979}\natexlab{}.
\newblock \showarticletitle{Scripts in memory for text}.
\newblock \bibinfo{journal}{\emph{Cognitive psychology}} \bibinfo{volume}{11},
  \bibinfo{number}{2} (\bibinfo{year}{1979}), \bibinfo{pages}{177--220}.
\newblock


\bibitem[Boyd et~al\mbox{.}(2020)]%
        {Boyd2020-jq}
\bibfield{author}{\bibinfo{person}{Ryan~L Boyd}, \bibinfo{person}{Kate~G
  Blackburn}, {and} \bibinfo{person}{James~W Pennebaker}.}
  \bibinfo{year}{2020}\natexlab{}.
\newblock \showarticletitle{The narrative arc: Revealing core narrative
  structures through text analysis}.
\newblock \bibinfo{journal}{\emph{Science advances}} \bibinfo{volume}{6},
  \bibinfo{number}{32} (\bibinfo{date}{Aug.} \bibinfo{year}{2020}),
  \bibinfo{pages}{eaba2196}.
\newblock


\bibitem[Brown et~al\mbox{.}(2020)]%
        {brown2020language}
\bibfield{author}{\bibinfo{person}{Tom~B Brown}, \bibinfo{person}{Benjamin
  Mann}, \bibinfo{person}{Nick Ryder}, \bibinfo{person}{Melanie Subbiah},
  \bibinfo{person}{Jared Kaplan}, \bibinfo{person}{Prafulla Dhariwal},
  \bibinfo{person}{Arvind Neelakantan}, \bibinfo{person}{Pranav Shyam},
  \bibinfo{person}{Girish Sastry}, \bibinfo{person}{Amanda Askell},
  {et~al\mbox{.}}} \bibinfo{year}{2020}\natexlab{}.
\newblock \bibinfo{title}{Language models are few-shot learners}.
  (\bibinfo{year}{2020}).
\newblock
\newblock
\shownote{unpublished}.


\bibitem[Brysbaert et~al\mbox{.}(2014)]%
        {brysbaert2014concreteness}
\bibfield{author}{\bibinfo{person}{Marc Brysbaert}, \bibinfo{person}{Amy~Beth
  Warriner}, {and} \bibinfo{person}{Victor Kuperman}.}
  \bibinfo{year}{2014}\natexlab{}.
\newblock \showarticletitle{Concreteness ratings for 40 thousand generally
  known {English} word lemmas}.
\newblock \bibinfo{journal}{\emph{Behavior Research Methods}}
  \bibinfo{volume}{46}, \bibinfo{number}{3} (\bibinfo{year}{2014}).
\newblock


\bibitem[Clark et~al\mbox{.}(2021)]%
        {clark-etal-2021-thats}
\bibfield{author}{\bibinfo{person}{Elizabeth Clark}, \bibinfo{person}{Tal
  August}, \bibinfo{person}{Sofia Serrano}, \bibinfo{person}{Nikita Haduong},
  \bibinfo{person}{Suchin Gururangan}, {and} \bibinfo{person}{Noah~A. Smith}.}
  \bibinfo{year}{2021}\natexlab{}.
\newblock \showarticletitle{All That{'}s {`}Human{'} Is Not Gold: Evaluating
  Human Evaluation of Generated Text}. In \bibinfo{booktitle}{\emph{Proceedings
  of the 59th Annual Meeting of the Association for Computational Linguistics
  and the 11th International Joint Conference on Natural Language Processing
  (Volume 1: Long Papers)}}. \bibinfo{publisher}{Association for Computational
  Linguistics}, \bibinfo{address}{Online}, \bibinfo{pages}{7282--7296}.
\newblock
\urldef\tempurl%
\url{https://doi.org/10.18653/v1/2021.acl-long.565}
\showDOI{\tempurl}


\bibitem[Clewett et~al\mbox{.}(2019)]%
        {clewett2019transcending}
\bibfield{author}{\bibinfo{person}{David Clewett}, \bibinfo{person}{Sarah
  DuBrow}, {and} \bibinfo{person}{Lila Davachi}.}
  \bibinfo{year}{2019}\natexlab{}.
\newblock \showarticletitle{Transcending time in the brain: How event memories
  are constructed from experience}.
\newblock \bibinfo{journal}{\emph{Hippocampus}} \bibinfo{volume}{29},
  \bibinfo{number}{3} (\bibinfo{year}{2019}), \bibinfo{pages}{162--183}.
\newblock


\bibitem[Conway et~al\mbox{.}(1996)]%
        {Conway1996-nx}
\bibfield{author}{\bibinfo{person}{Martin~A. Conway}, \bibinfo{person}{Alan~F.
  Collins}, \bibinfo{person}{Susan~E. Gathercole}, {and}
  \bibinfo{person}{Stephen~J. Anderson}.} \bibinfo{year}{1996}\natexlab{}.
\newblock \showarticletitle{Recollections of true and false autobiographical
  memories}.
\newblock \bibinfo{journal}{\emph{Journal of Experimental Psychology: General}}
  \bibinfo{volume}{125}, \bibinfo{number}{1} (\bibinfo{year}{1996}).
\newblock


\bibitem[Conway et~al\mbox{.}(2003)]%
        {Conway2003Neurophysiological}
\bibfield{author}{\bibinfo{person}{Martin~A. Conway},
  \bibinfo{person}{Christopher~W. Pleydell-Pearce}, \bibinfo{person}{Sharron~E.
  Whitecross}, {and} \bibinfo{person}{Helen Sharpe}.}
  \bibinfo{year}{2003}\natexlab{}.
\newblock \showarticletitle{Neurophysiological correlates of memory for
  experienced and imagined events}.
\newblock \bibinfo{journal}{\emph{Neuropsychologia}} \bibinfo{volume}{41},
  \bibinfo{number}{3} (\bibinfo{year}{2003}), \bibinfo{pages}{334--340}.
\newblock


\bibitem[Devlin et~al\mbox{.}(2019)]%
        {Devlin2019bert}
\bibfield{author}{\bibinfo{person}{Jacob Devlin}, \bibinfo{person}{Ming-Wei
  Chang}, \bibinfo{person}{Kenton Lee}, {and} \bibinfo{person}{Kristina
  Toutanova}.} \bibinfo{year}{2019}\natexlab{}.
\newblock \showarticletitle{{BERT}: Pre-training of Deep Bidirectional
  Transformers for Language Understanding}. In
  \bibinfo{booktitle}{\emph{NAACL}}.
\newblock
\urldef\tempurl%
\url{https://www.aclweb.org/anthology/N19-1423}
\showURL{%
\tempurl}


\bibitem[Franklin et~al\mbox{.}(2020)]%
        {franklin2020structured}
\bibfield{author}{\bibinfo{person}{Nicholas~T Franklin},
  \bibinfo{person}{Kenneth~A Norman}, \bibinfo{person}{Charan Ranganath},
  \bibinfo{person}{Jeffrey~M Zacks}, {and} \bibinfo{person}{Samuel~J
  Gershman}.} \bibinfo{year}{2020}\natexlab{}.
\newblock \showarticletitle{Structured Event Memory: A neuro-symbolic model of
  event cognition.}
\newblock \bibinfo{journal}{\emph{Psychological Review}} \bibinfo{volume}{127},
  \bibinfo{number}{3} (\bibinfo{year}{2020}), \bibinfo{pages}{327}.
\newblock


\bibitem[Gilboa et~al\mbox{.}(2018)]%
        {gilboa2018autobiographical}
\bibfield{author}{\bibinfo{person}{Asaf Gilboa}, \bibinfo{person}{R~Shayna
  Rosenbaum}, {and} \bibinfo{person}{Avi Mendelsohn}.}
  \bibinfo{year}{2018}\natexlab{}.
\newblock \showarticletitle{Autobiographical memory: From experiences to brain
  representations}.
\newblock \bibinfo{journal}{\emph{Neuropsychologia}}  \bibinfo{volume}{110}
  (\bibinfo{date}{Feb.} \bibinfo{year}{2018}), \bibinfo{pages}{1--6}.
\newblock


\bibitem[Goldstein et~al\mbox{.}(2021)]%
        {Goldstein2021thinking}
\bibfield{author}{\bibinfo{person}{Ariel Goldstein}, \bibinfo{person}{Zaid
  Zada}, \bibinfo{person}{Eliav Buchnik}, \bibinfo{person}{Mariano Schain},
  \bibinfo{person}{Amy Price}, \bibinfo{person}{Bobbi Aubrey},
  \bibinfo{person}{Samuel~A. Nastase}, \bibinfo{person}{Amir Feder},
  \bibinfo{person}{Dotan Emanuel}, \bibinfo{person}{Alon Cohen},
  \bibinfo{person}{Aren Jansen}, \bibinfo{person}{Harshvardhan Gazula},
  \bibinfo{person}{Gina Choe}, \bibinfo{person}{Aditi Rao},
  \bibinfo{person}{Catherine Kim}, \bibinfo{person}{Colton Casto},
  \bibinfo{person}{Lora Fanda}, \bibinfo{person}{Werner Doyle},
  \bibinfo{person}{Daniel Friedman}, \bibinfo{person}{Patricia Dugan},
  \bibinfo{person}{Roi Reichart}, \bibinfo{person}{Sasha Devore},
  \bibinfo{person}{Adeen Flinker}, \bibinfo{person}{Liat Hasenfratz},
  \bibinfo{person}{Avinatan Hassidim}, \bibinfo{person}{Michael Brenner},
  \bibinfo{person}{Yossi Matias}, \bibinfo{person}{Kenneth~A. Norman},
  \bibinfo{person}{Orrin Devinsky}, {and} \bibinfo{person}{Uri Hasson}.}
  \bibinfo{year}{2021}\natexlab{}.
\newblock \showarticletitle{Thinking ahead: spontaneous prediction in context
  as a keystone of language in humans and machines}. In
  \bibinfo{booktitle}{\emph{bioRxiv}}. \bibinfo{publisher}{Cold Spring Harbor
  Laboratory}.
\newblock
\urldef\tempurl%
\url{https://www.biorxiv.org/content/early/2021/03/19/2020.12.02.403477}
\showURL{%
\tempurl}


\bibitem[Gordon and Van~Durme(2013)]%
        {gordon2013reporting}
\bibfield{author}{\bibinfo{person}{Jonathan Gordon} {and}
  \bibinfo{person}{Benjamin Van~Durme}.} \bibinfo{year}{2013}\natexlab{}.
\newblock \showarticletitle{Reporting bias and knowledge acquisition}. In
  \bibinfo{booktitle}{\emph{Proceedings of the 2013 workshop on Automated
  knowledge base construction}}. \bibinfo{pages}{25--30}.
\newblock


\bibitem[Graesser et~al\mbox{.}(1981)]%
        {Graesser1981-uj}
\bibfield{author}{\bibinfo{person}{Arthur~C Graesser}, \bibinfo{person}{Scott~P
  Robertson}, {and} \bibinfo{person}{Patricia~A Anderson}.}
  \bibinfo{year}{1981}\natexlab{}.
\newblock \showarticletitle{Incorporating inferences in narrative
  representations: A study of how and why}.
\newblock \bibinfo{journal}{\emph{Cognitive Psychology}} \bibinfo{volume}{13},
  \bibinfo{number}{1} (\bibinfo{date}{Jan.} \bibinfo{year}{1981}),
  \bibinfo{pages}{1--26}.
\newblock


\bibitem[Greenberg et~al\mbox{.}(1996)]%
        {Greenberg1996-yt}
\bibfield{author}{\bibinfo{person}{Melanie~A. Greenberg},
  \bibinfo{person}{Camille~B. Wortman}, {and} \bibinfo{person}{Arthur~A.
  Stone}.} \bibinfo{year}{1996}\natexlab{}.
\newblock \showarticletitle{Emotional expression and physical health: revising
  traumatic memories or fostering self-regulation?}
\newblock \bibinfo{journal}{\emph{Journal of Personality and Social
  Psychology}} \bibinfo{volume}{71}, \bibinfo{number}{3} (\bibinfo{date}{Sept.}
  \bibinfo{year}{1996}), \bibinfo{pages}{588--602}.
\newblock


\bibitem[Grysman and Hudson(2011)]%
        {Grysman2011-jj}
\bibfield{author}{\bibinfo{person}{Azriel Grysman} {and}
  \bibinfo{person}{Judith~A Hudson}.} \bibinfo{year}{2011}\natexlab{}.
\newblock \showarticletitle{The self in autobiographical memory: effects of
  self-salience on narrative content and structure}.
\newblock \bibinfo{journal}{\emph{Memory}} \bibinfo{volume}{19},
  \bibinfo{number}{5} (\bibinfo{date}{July} \bibinfo{year}{2011}),
  \bibinfo{pages}{501--513}.
\newblock
\urldef\tempurl%
\url{http://dx.doi.org/10.1080/09658211.2011.590502}
\showURL{%
\tempurl}


\bibitem[Hale(2001)]%
        {hale2001probabilistic}
\bibfield{author}{\bibinfo{person}{John Hale}.}
  \bibinfo{year}{2001}\natexlab{}.
\newblock \showarticletitle{A probabilistic Earley parser as a psycholinguistic
  model}. In \bibinfo{booktitle}{\emph{{NAACL-HLT}}}. Association for
  Computational Linguistics, \bibinfo{pages}{1--8}.
\newblock


\bibitem[Hyman~Jr and Loftus(1998)]%
        {hyman1998errors}
\bibfield{author}{\bibinfo{person}{Ira~E Hyman~Jr} {and}
  \bibinfo{person}{Elizabeth~F Loftus}.} \bibinfo{year}{1998}\natexlab{}.
\newblock \showarticletitle{Errors in autobiographical memory}.
\newblock \bibinfo{journal}{\emph{Clinical psychology review}}
  \bibinfo{volume}{18}, \bibinfo{number}{8} (\bibinfo{year}{1998}),
  \bibinfo{pages}{933--947}.
\newblock


\bibitem[Jafarpour et~al\mbox{.}(2019a)]%
        {jafarpour2019event}
\bibfield{author}{\bibinfo{person}{Anna Jafarpour},
  \bibinfo{person}{Elizabeth~A Buffalo}, \bibinfo{person}{Robert~T Knight},
  {and} \bibinfo{person}{Anne~GE Collins}.} \bibinfo{year}{2019}\natexlab{a}.
\newblock \bibinfo{title}{Event segmentation reveals working memory forgetting
  rate}.
\newblock
\newblock


\bibitem[Jafarpour et~al\mbox{.}(2019b)]%
        {jafarpour2019medial}
\bibfield{author}{\bibinfo{person}{Anna Jafarpour}, \bibinfo{person}{Sandon
  Griffin}, \bibinfo{person}{Jack~J Lin}, {and} \bibinfo{person}{Robert~T
  Knight}.} \bibinfo{year}{2019}\natexlab{b}.
\newblock \showarticletitle{Medial orbitofrontal cortex, dorsolateral
  prefrontal cortex, and hippocampus differentially represent the event
  saliency}.
\newblock \bibinfo{journal}{\emph{Journal of cognitive neuroscience}}
  \bibinfo{volume}{31}, \bibinfo{number}{6} (\bibinfo{year}{2019}),
  \bibinfo{pages}{874--884}.
\newblock


\bibitem[Jaidka et~al\mbox{.}(2020)]%
        {Jaidka2020-gz}
\bibfield{author}{\bibinfo{person}{Kokil Jaidka}, \bibinfo{person}{Salvatore
  Giorgi}, \bibinfo{person}{H~Andrew Schwartz}, \bibinfo{person}{Margaret~L
  Kern}, \bibinfo{person}{Lyle~H Ungar}, {and} \bibinfo{person}{Johannes~C
  Eichstaedt}.} \bibinfo{year}{2020}\natexlab{}.
\newblock \showarticletitle{Estimating geographic subjective well-being from
  Twitter: A comparison of dictionary and data-driven language methods}.
\newblock \bibinfo{journal}{\emph{Proceedings of the National Academy of
  Sciences of the United States of America}} \bibinfo{volume}{117},
  \bibinfo{number}{19} (\bibinfo{date}{May} \bibinfo{year}{2020}),
  \bibinfo{pages}{10165--10171}.
\newblock
\urldef\tempurl%
\url{http://dx.doi.org/10.1073/pnas.1906364117}
\showURL{%
\tempurl}


\bibitem[Kintsch(1988)]%
        {kintsch1988role}
\bibfield{author}{\bibinfo{person}{Walter Kintsch}.}
  \bibinfo{year}{1988}\natexlab{}.
\newblock \showarticletitle{The role of knowledge in discourse comprehension: A
  construction-integration model.}
\newblock \bibinfo{journal}{\emph{Psychological Review}} \bibinfo{volume}{95},
  \bibinfo{number}{2} (\bibinfo{year}{1988}), \bibinfo{pages}{163}.
\newblock


\bibitem[Kintsch and Greene(1978)]%
        {kintsch1978role}
\bibfield{author}{\bibinfo{person}{Walter Kintsch} {and} \bibinfo{person}{Edith
  Greene}.} \bibinfo{year}{1978}\natexlab{}.
\newblock \showarticletitle{The role of culture-specific schemata in the
  comprehension and recall of stories}.
\newblock \bibinfo{journal}{\emph{Discourse processes}} \bibinfo{volume}{1},
  \bibinfo{number}{1} (\bibinfo{year}{1978}), \bibinfo{pages}{1--13}.
\newblock


\bibitem[{Kourken Michaelian}(2018)]%
        {Kourken_Michaelian2018-lg}
\bibfield{author}{\bibinfo{person}{{Kourken Michaelian}}.}
  \bibinfo{year}{2018}\natexlab{}.
\newblock \showarticletitle{Episodic and Semantic Memory and Imagination: The
  Need for Definitions}.
\newblock \bibinfo{journal}{\emph{The American journal of psychology}}
  \bibinfo{volume}{131}, \bibinfo{number}{1} (\bibinfo{year}{2018}),
  \bibinfo{pages}{99--103}.
\newblock
\urldef\tempurl%
\url{https://www.jstor.org/stable/10.5406/amerjpsyc.131.1.0099}
\showURL{%
\tempurl}


\bibitem[Kurby and Zacks(2008a)]%
        {Kurby2008-km}
\bibfield{author}{\bibinfo{person}{Christopher~A Kurby} {and}
  \bibinfo{person}{Jeffrey~M Zacks}.} \bibinfo{year}{2008}\natexlab{a}.
\newblock \showarticletitle{Segmentation in the perception and memory of
  events}.
\newblock \bibinfo{journal}{\emph{Trends in cognitive sciences}}
  \bibinfo{volume}{12}, \bibinfo{number}{2} (\bibinfo{date}{Feb.}
  \bibinfo{year}{2008}), \bibinfo{pages}{72--79}.
\newblock


\bibitem[Kurby and Zacks(2008b)]%
        {kurby2008segmentation}
\bibfield{author}{\bibinfo{person}{Christopher~A Kurby} {and}
  \bibinfo{person}{Jeffrey~M Zacks}.} \bibinfo{year}{2008}\natexlab{b}.
\newblock \showarticletitle{Segmentation in the perception and memory of
  events}.
\newblock \bibinfo{journal}{\emph{Trends in cognitive sciences}}
  \bibinfo{volume}{12}, \bibinfo{number}{2} (\bibinfo{year}{2008}),
  \bibinfo{pages}{72--79}.
\newblock


\bibitem[Laban et~al\mbox{.}(2021)]%
        {laban-etal-2021-transformer}
\bibfield{author}{\bibinfo{person}{Philippe Laban}, \bibinfo{person}{Luke Dai},
  \bibinfo{person}{Lucas Bandarkar}, {and} \bibinfo{person}{Marti~A. Hearst}.}
  \bibinfo{year}{2021}\natexlab{}.
\newblock \showarticletitle{Can Transformer Models Measure Coherence In Text:
  Re-Thinking the Shuffle Test}. In \bibinfo{booktitle}{\emph{Proceedings of
  the 59th Annual Meeting of the Association for Computational Linguistics and
  the 11th International Joint Conference on Natural Language Processing
  (Volume 2: Short Papers)}}. \bibinfo{publisher}{Association for Computational
  Linguistics}, \bibinfo{address}{Online}, \bibinfo{pages}{1058--1064}.
\newblock
\urldef\tempurl%
\url{https://doi.org/10.18653/v1/2021.acl-short.134}
\showDOI{\tempurl}


\bibitem[Levy(2008)]%
        {levy2008expectation}
\bibfield{author}{\bibinfo{person}{Roger Levy}.}
  \bibinfo{year}{2008}\natexlab{}.
\newblock \showarticletitle{Expectation-based syntactic comprehension}.
\newblock \bibinfo{journal}{\emph{Cognition}} \bibinfo{volume}{106},
  \bibinfo{number}{3} (\bibinfo{year}{2008}), \bibinfo{pages}{1126--1177}.
\newblock


\bibitem[Li et~al\mbox{.}(2013)]%
        {li2013story}
\bibfield{author}{\bibinfo{person}{Boyang Li}, \bibinfo{person}{Stephen
  Lee-Urban}, \bibinfo{person}{George Johnston}, {and} \bibinfo{person}{Mark
  Riedl}.} \bibinfo{year}{2013}\natexlab{}.
\newblock \showarticletitle{Story generation with crowdsourced plot graphs}. In
  \bibinfo{booktitle}{\emph{Twenty-Seventh AAAI Conference on Artificial
  Intelligence}}.
\newblock


\bibitem[Marsh(2007)]%
        {Marsh2007-ud}
\bibfield{author}{\bibinfo{person}{Elizabeth~J Marsh}.}
  \bibinfo{year}{2007}\natexlab{}.
\newblock \showarticletitle{Retelling Is Not the Same as Recalling:
  Implications for Memory}.
\newblock \bibinfo{journal}{\emph{Current directions in psychological science}}
  \bibinfo{volume}{16}, \bibinfo{number}{1} (\bibinfo{date}{Feb.}
  \bibinfo{year}{2007}), \bibinfo{pages}{16--20}.
\newblock
\urldef\tempurl%
\url{https://doi.org/10.1111/j.1467-8721.2007.00467.x}
\showURL{%
\tempurl}


\bibitem[Nabi and Green(2015)]%
        {nabi2015role}
\bibfield{author}{\bibinfo{person}{Robin~L Nabi} {and}
  \bibinfo{person}{Melanie~C Green}.} \bibinfo{year}{2015}\natexlab{}.
\newblock \showarticletitle{The role of a narrative's emotional flow in
  promoting persuasive outcomes}.
\newblock \bibinfo{journal}{\emph{Media Psychology}} \bibinfo{volume}{18},
  \bibinfo{number}{2} (\bibinfo{year}{2015}), \bibinfo{pages}{137--162}.
\newblock


\bibitem[Pennebaker et~al\mbox{.}(2015)]%
        {liwc2015}
\bibfield{author}{\bibinfo{person}{James~W. Pennebaker},
  \bibinfo{person}{Roger~J. Booth}, \bibinfo{person}{Ryan~L. Boyd}, {and}
  \bibinfo{person}{Martha~E. Francis}.} \bibinfo{year}{2015}\natexlab{}.
\newblock \bibinfo{title}{Linguistic Inquiry and Word Count: {LIWC} 2015}.
\newblock
\newblock


\bibitem[Piper et~al\mbox{.}(2021)]%
        {Piper2021-bj}
\bibfield{author}{\bibinfo{person}{Andrew Piper}, \bibinfo{person}{Richard~Jean
  So}, {and} \bibinfo{person}{David Bamman}.} \bibinfo{year}{2021}\natexlab{}.
\newblock \showarticletitle{Narrative Theory for Computational Narrative
  Understanding}. In \bibinfo{booktitle}{\emph{{EMNLP}}}.
  \bibinfo{publisher}{Association for Computational Linguistics},
  \bibinfo{address}{Online and Punta Cana, Dominican Republic},
  \bibinfo{pages}{298--311}.
\newblock
\urldef\tempurl%
\url{https://aclanthology.org/2021.emnlp-main.26}
\showURL{%
\tempurl}


\bibitem[Radford et~al\mbox{.}(2018)]%
        {Radford2018gpt}
\bibfield{author}{\bibinfo{person}{Alec Radford}, \bibinfo{person}{Karthik
  Narasimhan}, \bibinfo{person}{Tim Salimans}, {and} \bibinfo{person}{Ilya
  Sutskever}.} \bibinfo{year}{2018}\natexlab{}.
\newblock \bibinfo{title}{Improving Language Understanding by Generative
  Pre-Training}.  (\bibinfo{year}{2018}).
\newblock
\newblock
\shownote{unpublished}.


\bibitem[Radford et~al\mbox{.}(2019)]%
        {Radford2019gpt2}
\bibfield{author}{\bibinfo{person}{Alec Radford}, \bibinfo{person}{Jeffrey Wu},
  \bibinfo{person}{Rewon Child}, \bibinfo{person}{David Luan},
  \bibinfo{person}{Dario Amodei}, {and} \bibinfo{person}{Ilya Sutskever}.}
  \bibinfo{year}{2019}\natexlab{}.
\newblock \bibinfo{title}{Language Models are Unsupervised Multitask Learners}.
   (\bibinfo{year}{2019}).
\newblock
\newblock
\shownote{unpublished}.


\bibitem[Reagan et~al\mbox{.}(2016)]%
        {reagan2016emotional}
\bibfield{author}{\bibinfo{person}{Andrew~J Reagan}, \bibinfo{person}{Lewis
  Mitchell}, \bibinfo{person}{Dilan Kiley}, \bibinfo{person}{Christopher~M
  Danforth}, {and} \bibinfo{person}{Peter~Sheridan Dodds}.}
  \bibinfo{year}{2016}\natexlab{}.
\newblock \showarticletitle{The emotional arcs of stories are dominated by six
  basic shapes}.
\newblock \bibinfo{journal}{\emph{EPJ Data Science}} \bibinfo{volume}{5},
  \bibinfo{number}{1} (\bibinfo{year}{2016}), \bibinfo{pages}{31}.
\newblock


\bibitem[Reichardt et~al\mbox{.}(2020)]%
        {reichardt2020novelty}
\bibfield{author}{\bibinfo{person}{Rich{\'a}rd Reichardt},
  \bibinfo{person}{Bertalan Polner}, {and} \bibinfo{person}{P{\'e}ter Simor}.}
  \bibinfo{year}{2020}\natexlab{}.
\newblock \showarticletitle{Novelty manipulations, memory performance, and
  predictive coding: The role of unexpectedness}.
\newblock \bibinfo{journal}{\emph{Frontiers in human neuroscience}}
  \bibinfo{volume}{14} (\bibinfo{year}{2020}), \bibinfo{pages}{152}.
\newblock


\bibitem[Sap et~al\mbox{.}(2020a)]%
        {sap-horvitz-etal-2020-recall-vs-imagined}
\bibfield{author}{\bibinfo{person}{Maarten Sap}, \bibinfo{person}{Eric
  Horvitz}, \bibinfo{person}{Yejin Choi}, \bibinfo{person}{Noah~A. Smith},
  {and} \bibinfo{person}{James~W. Pennebaker}.}
  \bibinfo{year}{2020}\natexlab{a}.
\newblock \showarticletitle{Recollection versus Imagination: Exploring Human
  Memory and Cognition via Neural Language Models,}. In
  \bibinfo{booktitle}{\emph{Proceedings of the Association for Computational
  Linguistics}}. \bibinfo{publisher}{Association for Computational
  Linguistics}, \bibinfo{address}{Seattle, Washington}.
\newblock


\bibitem[Sap et~al\mbox{.}(2020b)]%
        {sap2020recollectionImagination}
\bibfield{author}{\bibinfo{person}{Maarten Sap}, \bibinfo{person}{Eric
  Horvitz}, \bibinfo{person}{Yejin Choi}, \bibinfo{person}{Noah~A Smith}, {and}
  \bibinfo{person}{James~W Pennebaker}.} \bibinfo{year}{2020}\natexlab{b}.
\newblock \showarticletitle{Recollection versus Imagination: Exploring Human
  Memory and Cognition via Neural Language Models}. In
  \bibinfo{booktitle}{\emph{ACL}}.
\newblock


\bibitem[Schacter and Addis(2007)]%
        {Schacter2007-as}
\bibfield{author}{\bibinfo{person}{Daniel~L Schacter} {and}
  \bibinfo{person}{Donna~Rose Addis}.} \bibinfo{year}{2007}\natexlab{}.
\newblock \showarticletitle{The cognitive neuroscience of constructive memory:
  remembering the past and imagining the future}.
\newblock \bibinfo{journal}{\emph{Philosophical Transactions of the Royal
  Society B: Biological Sciences}} \bibinfo{volume}{362},
  \bibinfo{number}{1481} (\bibinfo{year}{2007}), \bibinfo{pages}{773--786}.
\newblock


\bibitem[Schank and Abelson(1977)]%
        {schank1977scripts}
\bibfield{author}{\bibinfo{person}{Roger~C. Schank} {and}
  \bibinfo{person}{Robert~P. Abelson}.} \bibinfo{year}{1977}\natexlab{}.
\newblock \bibinfo{booktitle}{\emph{Scripts, Plans, Goals and Understanding: An
  Inquiry into Human Knowledge Structures}}.
\newblock \bibinfo{publisher}{Lawrence Erlbaum}.
\newblock


\bibitem[Sims et~al\mbox{.}(2019)]%
        {Sims2019Literary}
\bibfield{author}{\bibinfo{person}{Matthew Sims}, \bibinfo{person}{Jong~Ho
  Park}, {and} \bibinfo{person}{David Bamman}.}
  \bibinfo{year}{2019}\natexlab{}.
\newblock \showarticletitle{Literary Event Detection}. In
  \bibinfo{booktitle}{\emph{{ACL}}}.
\newblock
\urldef\tempurl%
\url{https://www.aclweb.org/anthology/P19-1353}
\showURL{%
\tempurl}


\bibitem[Smorti and Fioretti(2016)]%
        {smorti2016narrating}
\bibfield{author}{\bibinfo{person}{Andrea Smorti} {and} \bibinfo{person}{Chiara
  Fioretti}.} \bibinfo{year}{2016}\natexlab{}.
\newblock \showarticletitle{Why narrating changes memory: a contribution to an
  integrative model of memory and narrative processes}.
\newblock \bibinfo{journal}{\emph{Integrative Psychological and Behavioral
  Science}} \bibinfo{volume}{50}, \bibinfo{number}{2} (\bibinfo{year}{2016}),
  \bibinfo{pages}{296--319}.
\newblock


\bibitem[Squire(1981)]%
        {squire1981two}
\bibfield{author}{\bibinfo{person}{LARRY~R Squire}.}
  \bibinfo{year}{1981}\natexlab{}.
\newblock \showarticletitle{Two forms of human amnesia: An analysis of
  forgetting}.
\newblock \bibinfo{journal}{\emph{Journal of Neuroscience}}
  \bibinfo{volume}{1}, \bibinfo{number}{6} (\bibinfo{year}{1981}),
  \bibinfo{pages}{635--640}.
\newblock


\bibitem[Tausczik and Pennebaker(2010)]%
        {tausczik2010psychological}
\bibfield{author}{\bibinfo{person}{Yla~R. Tausczik} {and}
  \bibinfo{person}{James~W. Pennebaker}.} \bibinfo{year}{2010}\natexlab{}.
\newblock \showarticletitle{The psychological meaning of words: LIWC and
  computerized text analysis methods}.
\newblock \bibinfo{journal}{\emph{Journal of Language and Social Psychology}}
  \bibinfo{volume}{29}, \bibinfo{number}{1} (\bibinfo{year}{2010}),
  \bibinfo{pages}{24--54}.
\newblock


\bibitem[Thorndyke(1977)]%
        {thorndyke1977cognitive}
\bibfield{author}{\bibinfo{person}{Perry~W Thorndyke}.}
  \bibinfo{year}{1977}\natexlab{}.
\newblock \showarticletitle{Cognitive structures in comprehension and memory of
  narrative discourse}.
\newblock \bibinfo{journal}{\emph{Cognitive psychology}} \bibinfo{volume}{9},
  \bibinfo{number}{1} (\bibinfo{year}{1977}), \bibinfo{pages}{77--110}.
\newblock


\bibitem[Toubia et~al\mbox{.}(2021)]%
        {Toubia2021-fm}
\bibfield{author}{\bibinfo{person}{Olivier Toubia}, \bibinfo{person}{Jonah
  Berger}, {and} \bibinfo{person}{Jehoshua Eliashberg}.}
  \bibinfo{year}{2021}\natexlab{}.
\newblock \showarticletitle{How quantifying the shape of stories predicts their
  success}.
\newblock \bibinfo{journal}{\emph{Proceedings of the National Academy of
  Sciences of the United States of America}} \bibinfo{volume}{118},
  \bibinfo{number}{26} (\bibinfo{date}{June} \bibinfo{year}{2021}).
\newblock


\bibitem[Tulving(1972)]%
        {tulving1972episodic}
\bibfield{author}{\bibinfo{person}{Endel Tulving}.}
  \bibinfo{year}{1972}\natexlab{}.
\newblock \showarticletitle{Episodic and semantic memory}.
\newblock \bibinfo{journal}{\emph{Organization of Memory}}  \bibinfo{volume}{1}
  (\bibinfo{year}{1972}), \bibinfo{pages}{381--403}.
\newblock


\bibitem[Underwood(2013)]%
        {underwood2013}
\bibfield{author}{\bibinfo{person}{Ted Underwood}.}
  \bibinfo{year}{2013}\natexlab{}.
\newblock \bibinfo{booktitle}{\emph{The Invention of Historical Perspective}}.
\newblock \bibinfo{publisher}{Stanford University Press},
  \bibinfo{pages}{55--80}.
\newblock


\bibitem[Underwood(2020)]%
        {underwood2020}
\bibfield{author}{\bibinfo{person}{Ted Underwood}.}
  \bibinfo{year}{2020}\natexlab{}.
\newblock \showarticletitle{Machine Learning and Human Perspective}.
\newblock \bibinfo{journal}{\emph{PMLA/Publications of the Modern Language
  Association of America}} \bibinfo{volume}{135}, \bibinfo{number}{1}
  (\bibinfo{year}{2020}), \bibinfo{pages}{92--109}.
\newblock


\bibitem[van Kesteren et~al\mbox{.}(2012)]%
        {Van_Kesteren2012-el}
\bibfield{author}{\bibinfo{person}{Marlieke T.~R. van Kesteren},
  \bibinfo{person}{Dirk~J. Ruiter}, \bibinfo{person}{Guill{\'e}n
  Fern{\'a}ndez}, {and} \bibinfo{person}{Richard~N. Henson}.}
  \bibinfo{year}{2012}\natexlab{}.
\newblock \showarticletitle{How schema and novelty augment memory formation}.
\newblock \bibinfo{journal}{\emph{Trends in Neurosciences}}
  \bibinfo{volume}{35}, \bibinfo{number}{4} (\bibinfo{date}{April}
  \bibinfo{year}{2012}).
\newblock


\bibitem[van Laer et~al\mbox{.}(2019)]%
        {Van_Laer2019-qq}
\bibfield{author}{\bibinfo{person}{Tom van Laer}, \bibinfo{person}{Jennifer
  Edson~Escalas}, \bibinfo{person}{Stephan Ludwig}, {and}
  \bibinfo{person}{Ellis~A van~den Hende}.} \bibinfo{year}{2019}\natexlab{}.
\newblock \showarticletitle{What Happens in Vegas Stays on {TripAdvisor}? A
  Theory and Technique to Understand Narrativity in Consumer Reviews}.
\newblock \bibinfo{journal}{\emph{The Journal of consumer research}}
  \bibinfo{volume}{46}, \bibinfo{number}{2} (\bibinfo{date}{Aug.}
  \bibinfo{year}{2019}), \bibinfo{pages}{267--285}.
\newblock
\urldef\tempurl%
\url{https://academic-oup-com.offcampus.lib.washington.edu/jcr/article-pdf/46/2/267/28916270/ucy067.pdf}
\showURL{%
\tempurl}


\bibitem[Zacks et~al\mbox{.}(2007)]%
        {zacks2007event}
\bibfield{author}{\bibinfo{person}{Jeffrey~M Zacks}, \bibinfo{person}{Nicole~K
  Speer}, \bibinfo{person}{Khena~M Swallow}, \bibinfo{person}{Todd~S Braver},
  {and} \bibinfo{person}{Jeremy~R Reynolds}.} \bibinfo{year}{2007}\natexlab{}.
\newblock \showarticletitle{Event perception: a mind-brain perspective.}
\newblock \bibinfo{journal}{\emph{Psychological bulletin}}
  \bibinfo{volume}{133}, \bibinfo{number}{2} (\bibinfo{year}{2007}),
  \bibinfo{pages}{273}.
\newblock


\bibitem[Zwaan and Radvansky(1998)]%
        {zwaan1998situation}
\bibfield{author}{\bibinfo{person}{Rolf~A Zwaan} {and}
  \bibinfo{person}{Gabriel~A Radvansky}.} \bibinfo{year}{1998}\natexlab{}.
\newblock \showarticletitle{Situation models in language comprehension and
  memory.}
\newblock \bibinfo{journal}{\emph{Psychological bulletin}}
  \bibinfo{volume}{123}, \bibinfo{number}{2} (\bibinfo{year}{1998}),
  \bibinfo{pages}{162}.
\newblock


\end{thebibliography}
